\documentclass[final,3p,times, authoryear]{elsarticle}
\usepackage{amssymb}
\usepackage{times}
\usepackage{graphicx}
\usepackage{subcaption}
\usepackage{amsmath}
\usepackage{amssymb}
\usepackage{url}
\usepackage{setspace}
\usepackage{titlesec}
\usepackage{algorithm}
\usepackage{algpseudocode}
\usepackage{listings}
\usepackage{amsmath,amssymb,amsthm}

\usepackage{hyperref}

\usepackage[dvipsnames]{xcolor}

\usepackage{tikz}
\usetikzlibrary{arrows.meta, positioning, trees}

\newcommand{\x}{{\mathbf x}}

\newcommand{\y}{{\mathbf y}}

\newcommand{\z}{{\mathbf z}}

\newcommand{\argmax}{\text{argmax}}

\graphicspath{
    {fig_new/}
    {img/}
    {img/introduction/}
    {img/gmm_test_cases/}
    {img/experimental_test_case/}
}

\journal{Computers \& Chemical Engineering}

\begin{document}

\begin{frontmatter}

%% Title, authors and addresses

%% use the tnoteref command within \title for footnotes;
%% use the tnotetext command for theassociated footnote;
%% use the fnref command within \author or \address for footnotes;
%% use the fntext command for theassociated footnote;
%% use the corref command within \author for corresponding author footnotes;
%% use the cortext command for theassociated footnote;
%% use the ead command for the email address,
%% and the form \ead[url] for the home page:
%% \title{Title\tnoteref{label1}}
%% \tnotetext[label1]{}
%% \author{Name\corref{cor1}\fnref{label2}}
%% \ead{email address}
%% \ead[url]{home page}
%% \fntext[label2]{}
%% \cortext[cor1]{}
%% \affiliation{organization={},
%%             addressline={},
%%             city={},
%%             postcode={},
%%             state={},
%%             country={}}
%% \fntext[label3]{}

\title{Process-constrained batch Bayesian approaches \\
for yield optimization in multi-reactor systems}

%% use optional labels to link authors explicitly to addresses:
%% \author[label1,label2]{}
%% \affiliation[label1]{organization={},
%%             addressline={},
%%             city={},
%%             postcode={},
%%             state={},
%%             country={}}
%%
%% \affiliation[label2]{organization={},
%%             addressline={},
%%             city={},
%%             postcode={},
%%             state={},
%%             country={}}

% \author[inst1]{Author One}

% \affiliation[inst1]{organization={Department One},%Department and Organization
%             addressline={Address One}, 
%             city={City One},
%             postcode={00000}, 
%             state={State One},
%             country={Country One}}

% \author[inst2]{Author Two}
% \author[inst1,inst2]{Author Three}

% \affiliation[inst2]{organization={Department Two},%Department and Organization
%             addressline={Address Two}, 
%             city={City Two},
%             postcode={22222}, 
%             state={State Two},
%             country={Country Two}}

% \begin{abstract}
% %% Text of abstract
% Lorem ipsum dolor sit amet, consectetur adipiscing elit, sed do eiusmod tempor incididunt ut labore et dolore magna aliqua. Ut enim ad minim veniam, quis nostrud exercitation ullamco laboris nisi ut aliquip ex ea commodo consequat. Duis aute irure dolor in reprehenderit in voluptate velit esse cillum dolore eu fugiat nulla pariatur. Excepteur sint occaecat cupidatat non proident, sunt in culpa qui officia deserunt mollit anim id est laborum.
% \end{abstract}

\author[inst1,inst2]{Markus Grimm}

\author[inst2]{Sébastien Paul}
\author[inst1]{Pierre Chainais}

\affiliation[inst1]{organization={Univ. Lille, CNRS, Centrale Lille},%Department and Organization
            addressline={UMR 9189 CRIStAL, F-59000 Lille}, 
            country={France}}

% 1. Univ. Lille, CNRS, Centrale Lille, UMR 9189 CRIStAL, F-59000 Lille, France

\affiliation[inst2]{
            organization={Univ. Lille, CNRS, Centrale Lille, Univ. Artois},%Department and Organization
            addressline={UMR 8181 - UCCS - Unite de Catalyse et Chimie du Solide, F-59000 Lille},
            country={France}}
% Univ. Lille, CNRS, Centrale Lille, 
% Univ. Artois, UMR 8181 - UCCS - 
% Unité de Catalyse et Chimie du Solide, 
% F-59000 Lille, France.

\begin{abstract}
%% Text of abstract
The optimization of yields in multi-reactor systems, which are advanced tools in heterogeneous catalysis research, presents a significant challenge due to hierarchical technical constraints. To this respect, this work introduces a novel approach called process-constrained batch Bayesian optimization via Thompson sampling (pc-BO-TS) and its generalized hierarchical extension (hpc-BO-TS). This method, tailored for the efficiency demands in multi-reactor systems, integrates experimental constraints and balances between exploration and exploitation in a sequential batch optimization strategy. It offers an improvement over other Bayesian optimization methods. The performance of pc-BO-TS and hpc-BO-TS is validated in synthetic cases as well as in a realistic scenario based on data obtained from high-throughput experiments done on a multi-reactor system available in the REALCAT platform. The proposed methods often outperform other sequential Bayesian optimizations and existing process-constrained batch Bayesian optimization methods. This work proposes a novel approach to optimize the yield of a reaction in a multi-reactor system, marking a significant step forward in digital catalysis and generally in optimization methods for chemical engineering.

\end{abstract}

% %%Graphical abstract
% \begin{graphicalabstract}
% \includegraphics{grabs}
% \end{graphicalabstract}

%%Research highlights
% \begin{highlights}
% \item Research highlight 1
% \item Research highlight 2
% \end{highlights}

\begin{keyword}
% % %% keywords here, in the form: keyword \sep keyword
% % keyword one \sep keyword two
% % %% PACS codes here, in the form: \PACS code \sep code
% % \PACS 0000 \sep 1111
% % %% MSC codes here, in the form: \MSC code \sep code
% % %% or \MSC[2008] code \sep code (2000 is the default)
% % \MSC 0000 \sep 1111
Process-constrained batch optimization \sep
Bayesian optimization \sep
Thompson sampling \sep
multi-reactor systems \sep
digital catalysis

\end{keyword}

\end{frontmatter}

%\tableofcontents

%%%%%%%%%%%%%%%%%%%%%%%%%%%%%%%%%%%   CURRENT INTRODUCTION
\section{Introduction}
\label{sec:Introduction}

%% General pb of optim reaction conditions in chem eng.
%
% Expensive => parallel => MRS 
Optimizing the conditions of a reaction, for instance, with respect to the yield of the target product, presents a significant challenge in chemical engineering. Sequential optimization using one single reactor typically requires extensive experimental time and resources. Multi-reactor systems (MRS) are key to High-Throughput Experiments (HTE). By enabling simultaneous exploration of the design space, they accelerate the optimization of the set of reaction conditions through parallel processing. %{\color{magenta} (what are the most general contexts?)} 
%% Parallel reactors
However, this parallel optimization process is intricately linked with constraints arising from the technical configuration of the multi-reactor system such as, for instance, a common feed composition or pressure for all reactors. These constraints create a layered hierarchy of parameters. The existence of this hierarchy of constraints leads to the emergence of process-constrained batch optimization problems.

\begin{figure}%[h!]
    \centering
    \includegraphics[scale=.35]{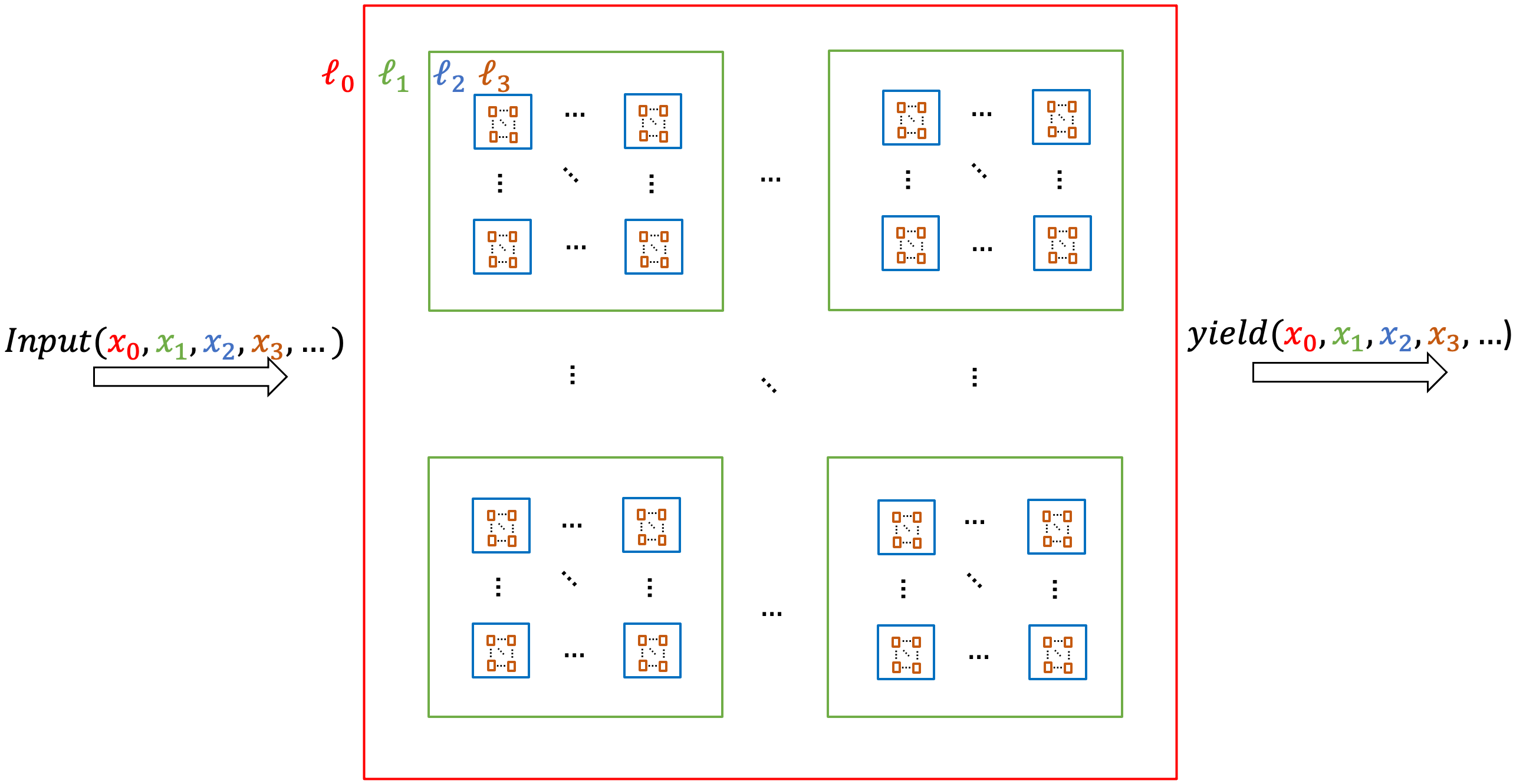}
    \caption{A hierarchical process-constrained (hpc) framework within a Multi-Reactor System, showcasing levels \( \ell_0 \) to \( \ell_{N-1} \) each with distinct batch sizes \( B_i \) and process constraints. Parameters \( x_k \) at each level are experimental conditions, with higher levels enforcing stricter constraints. 
    %This structure allows control over experimental variables with hierarchical constraints to optimize yield, reflecting the concept of hpc in HTE platforms. 
    }
    \label{fig:hierarchy_scheme}
\end{figure}

%% Fig. 1 : hierarchy of constraints in MRS
Figure~\ref{fig:hierarchy_scheme} illustrates the complexity and flexibility of the hierarchical process-constrained approach in accommodating various operational constraints across $N$ multiple levels $0\leq \ell\leq N-1$ of the experimental setup. Each level $\ell$ of the hierarchy can accommodate a distinct set of experimental parameters, reflecting the nested and recursive nature of constraints and degrees of freedom within the MRS. 
Each level $\ell_i$ of process constraints comes with a varying batch size $B_i$ of sub-reactor systems. The corresponding degrees of freedom are all the parameters $x_i$ but also all the $x_j$'s for $j\geq i$ at lower levels $\ell_j$. 
Note that $x_i$ will be part of constrained parameters for subsequent levels of the hierarchy.
At the highest level, the constraints are the most stringent, while lower levels allow for increasing parameter variability within each block. This principle enables a hierarchical control of constrained experimental parameters, resulting in the target yield of the process, as captured by the yield function \( yield(x_1, x_2, x_3, x_4, \ldots, x_d)\). 

%% A simple case: Flowrence within REALCAT 
%% REALCAT
The REALCAT's Flowrence unit from Avantium features a continuous catalytic MRS which fits this context. It features a hierarchy of fixed-bed reactors divided into 4 blocks of 4 reactors, in an architecture similar to Figure \ref{fig:hierarchy_scheme} \citep{Paul2015}. These reactors are all fed with an identical reactant flow. Each block itself permits an independent temperature control, that is common to 4 reactors. Each of the 4 reactors is loaded with a different catalyst mass. This setup illustrates the complexity of the optimization of the reaction parameters that maximize the yield under such process constraints. There is a crucial need for specialized optimization approaches.

%\pc{Modif: (from sec. 2.3)}
In heterogeneous catalysis, Bayesian optimization (BO) has accelerated process modeling and optimization, thus enhancing both simulated and real catalytic reactions \cite{YAN2011}. Its application extends to catalyst development and materials design, as shown in several studies \cite{Ohyama22, Frazier2016, WANG2022100728, WALKER2022100820}, highlighting its versatility and efficiency. This reinforces the interest of Bayesian sequential optimization approaches, suited for navigating complex, multi-dimensional optimization landscapes.
In general, the utility of BO is evident in scenarios involving complex simulations or experimental processes where evaluations of the objective function are expensive. Its effectiveness has been demonstrated across various domains, such as chemical engineering. For instance, Lu et al. \cite{LU2021107491} successfully employed sequential BO for optimizing a model predictive control system in HVAC plants. 
%
%% Acquisition functions and TS
BO relies on some predictive surrogate of the objective function, called the {\em acquisition function}, see section~\ref{subsec:BO} for a detailed presentation. Commonly used acquisition functions are GP-UCB (Gaussian Process Upper Confidence Bound) \cite{Srinivas2012_GPUCB}, the Expected Improvement (EI), or the Probability of Improvement (PI) \cite{Shahriari2016}.
Thompson Sampling (TS) is another acquisition function that has been effectively integrated into BO frameworks with Gaussian Processes, as demonstrated by \cite{Kandasamy2018}. In particular, Thompson sampling is a stochastic approach that randomly samples the surrogate objective function from a posterior distribution knowing the results of previous experiments. This work proposes to take benefit from the good exploration properties of the optimization domain by TS.

%% SOTA of Optim methods
The current state-of-the-art methods, including sequential and parallel Bayesian optimization (BO), can be successfully applied to problems in material science and hyperparameter tuning \citep{Frazier2016, FRANCESCHINI_2008, NEURIPS2020_BO_hyper}. 
%% Vellanki and process constrained
However, they are not fully equipped to address the constraints of multi-reactor 
systems.
Works like \citet{Vellanki2017} have developed techniques for process-constrained batch optimization. However, these methods and other constrained BO approaches \citep{Gelbart_2014_unkown_constraints, Gardner_2014_inequality_constraints} lack efficiency and struggle with the complex constraints inherent to multi-reactor setups.

%%%%% FIGURE Hierarchy
%%%%

%% In the spririt of Vellanki, a new method for a hopefully better trade-off between exploration and exploitation: our pc-BO-TS  [no math guarantees, but empirical performance will result ok]
This article proposes a process-constrained batch Bayesian optimization via Thompson sampling (pc-BO-TS) and its hierarchical extension (hpc-BO-TS) to deep multi-reactor systems. This approach demonstrates improved empirical performance by effectively balancing exploration and exploitation using Thompson Sampling for determining constrained parameters.
%
%% Empirical performance proof of concept: a set of numerical experiments: 1) synthetic cases, 2) a realistic test case based on a real experimental setup and data cf. REALCAT.
Numerical experiments permit us to illustrate and compare the empirical performance of various methods. In synthetic test cases, the proposed method favorably compares to state-of-the-art on a range of low to higher dimensional problems. % Revision 2 Comment 1
In particular, the pc-BO-TS approach exhibits improved performance, combined with either the EI or UCB acquisition functions. This indicates that the approach is robust, and independent of the chosen acquisition function.
%where the performance is comparable with a slight advantage for our proposed method. 
In a realistic test case based on the HTE REALCAT platform \citep{Paul2015}, pc-BO-TS empirically surpasses existing approaches and also expands the set of process-constrained batch BO techniques. 
This result showcases how such methods can be pragmatically applied in multi-reactor systems to optimize heterogeneously catalyzed reactions. Thereby, this finding accompanies the very recent shift of this field towards more data-centric methods known as {\em Digital Catalysis} \citep{digitalCatalysisMarshall}.

%% Claiming contributions
In summary, the contributions of this paper are: 
\begin{itemize}
    \item the introduction of a novel optimization approach called process-constrained Bayesian optimization using Thompson sampling (pc-BO-TS), offering a new alternative to existing methods in the field;
    \item the hierarchical extension (hpc-BO-TS) of the proposed approach to deep multi-level problems;
    \item the demonstration of enhanced empirical performance in relevant scenarios through the effective balance of exploration and exploitation, particularly in determining unconstrained parameters;
    \item a practical application to a realistic multi-reactor system for a successful validation and application of this method, particularly to optimize heterogeneously catalyzed reactions in multi-reactor systems, contributing to the advancement of digital catalysis.
\end{itemize}

% Structure of the Paper
The remainder of this article is organized as follows. 
Section~\ref{sec:seqOptimization} discusses the fundamentals of sequential BO, 
detailing its applications and relevance to our study. It also introduces 
parallel Bayesian optimization and defines our specific challenge in process-constrained 
batch optimization. 
Section~\ref{sec:pcBO} outlines the proposed methods, elaborating on their 
design and implementation. 
Section~\ref{sec:syntheticTestCase} introduces 
synthetic test cases used for evaluating our approach, providing a 
controlled assessment of its performance. 
Section~\ref{sec:realistic_case_study} applies 
our method to realistic data scenarios, demonstrating its practical effectiveness. 
Finally, Section~\ref{sec:conclusion} concludes the paper with a summary of our 
findings and a discussion on future directions for research.

%\clearpage

%%%%%%%%%%%%%%%%%%%%%%%%%%%%%%%
\section{From sequential to parallel Bayesian optimization}
\label{sec:seqOptimization}

Optimization problems in engineering and applied sciences frequently involve 'black box' objective functions. This function, here denoted as \( f: \mathcal{X} \subset \mathbb{R}^n \to \mathbb{R} \), lacks an analytical form \citep{Shahriari2016} and it can correspond to a real experiment. Such functions are often expensive to evaluate, making extensive sampling impractical.
The primary goal is to identify the optimal value \( x^* \) that maximizes or minimizes \( f(x) \) over the domain of interest \( \mathcal{X} \). Mathematically, this is expressed as:
\begin{equation}
    x^* = \arg\max_{x \in \mathcal{X}} f(x),
\label{eq:globOptimization} 
\end{equation}
There are hosts of approaches to solve such problems. This section focuses on Bayesian approaches based on Gaussian processes to give the background necessary for the introduction of the parallel process-constrained batch optimization in the next section.

%%%
\subsection{Sequential optimization}
\label{subsec:seq-optim}

Sequential optimization usually builds and iteratively updates a surrogate model \( \tilde{f}(x; \theta_t) \), parameterized by \( \theta_t \) at iteration \( t \). This approximate model of $f$ is iteratively updated by using outputs collected from previous evaluations. The selection of \( x_t \) at each iteration \( t \) is guided by this model, aiming to efficiently explore the search space using a sequence of proposals:
\begin{equation}
    x_t = \arg\max_{x \in \mathcal{X}} \tilde{f}(x; \theta_t).
\label{eq:seqOptimization}
\end{equation}
The concept of cumulative regret \( R_T \) is often used to quantify the opportunity loss from not selecting the optimal set of parameters over a set of iterations. It is mathematically defined as:
\begin{equation}
    R_T = \sum_{t=1}^{T} \left| f(x^*) - f(x_t) \right|,
\label{eq:cumulativeRegret}
\end{equation}
where \( T \) represents the total number of iterations (the budget), and \( t \) is the current iteration. Minimizing \( R_T \) is a usual metric of sequential optimization and depends significantly on the strategy for selecting \( x_t \).
The cumulative regret is relevant in cases where the degree of approximation of intermediate steps generates some loss.
% Normalized regret
When only the position and value of the optimum estimated after a finite number of iterations are at stake, the normalized regret is more suitable. 
% Positive functions ranging from 0 to maximum.
For positive functions ranging from 0 to their maximum, it is defined in a proper manner as:
\begin{equation}
    \label{eq:normalized-risk}
    R_{t} = 1 - \frac{f(x_t)}{f(x^*)}.
\end{equation}
The normalized regret ranges from 0 to 1. A normalized regret of 0 indicates a perfect identification of the optimum while 1 corresponds to $f(x_t)=0$, that is when $x_t$ corresponds to the minimum of $f$ (worst case scenario).
The distribution of normalized regrets for a variety of initial conditions permits a detailed analysis of the performance in terms of accuracy as well as robustness: a tighter distribution translates to a smaller dependency on initial conditions, and therefore a higher robustness. 
These metrics will be used in sections~\ref{sec:syntheticTestCase} and \ref{sec:realistic_case_study} to study the performance of the proposed methods compared to the literature. 
%findings based on these statistics are discussed in detail in Subsection \ref{subsubsec:synthetic_results_and_findings}.

%%%%%

%The strategy chosen to update \( \mathbf{x}_t \) from the data accumulated from previous iterations is called a policy. 
Various testing strategies can be employed based on the surrogate model's current understanding of the objective function. They all rely on uncertainty reduction as well as on an exploration-exploitation trade-off.
Sequential optimization is particularly valuable in fields like chemical engineering, where experiments and simulations are resource-intensive. For example, optimizing reaction conditions on the MRS Flowrence unit of REALCAT with a finite resource budget can be modeled as a sequential optimization problem as described in eq.~\eqref{eq:seqOptimization}.

\subsection{Gaussian Processes}
\label{subsec:GP}
%In the context of modeling an unknown objective function \(f\), 
Gaussian Processes (GPs) offer a non-parametric statistical modeling approach suitable for complex or poorly understood functions \citep{Rasmussen2004}. GPs permit us to define a distribution of surrogate functions \(\tilde{f}\). In practice, they are defined as a collection of jointly Gaussian random variables. This representation is formalized as \(\tilde{f} \sim \mathcal{GP}(\mu_{\theta}(\cdot), k_{\theta}(\cdot, \cdot))\), where \(\mu_{\theta} : \mathcal{X} \to \mathbb{R}\) is the mean function, and \(k_{\theta} : \mathcal{X} \times \mathcal{X} \to \mathbb{R}\) is the covariance function or kernel. The kernel hyperparameters, denoted by \(\theta\), depend on the choice of the kernel, such as the RBF or Matérn kernel. A common hyperparameter for all kernels is the length scale. The hyperparameters \(\theta\) play a crucial role in determining the characteristics of the covariance function, such as its smoothness which in turn, indirectly influences the mean function.

Given a dataset of \(n\) points \(x_{1:n}\), with corresponding function values \(\boldsymbol{\varphi} = (\varphi_i)_{1:n}\) and noisy observations \(\boldsymbol{y} = y_{1:n}\), GP regression models these as jointly Gaussian. The mean vector \(\boldsymbol{m}_{\theta}\) and the covariance matrix \(\boldsymbol{K}_{\theta}\), derived from the mean and covariance functions, incorporate the kernel hyperparameters \(\theta\):
\begin{eqnarray}
    \boldsymbol{\varphi} | \boldsymbol{X}, \theta & \sim &
    \mathcal{N}(\boldsymbol{m}_{\theta}(\boldsymbol{X}), \boldsymbol{K}_{\theta}(\boldsymbol{X}, \boldsymbol{X})),\\
%\end{equation}
%\begin{equation}
    \boldsymbol{y} | \boldsymbol{\varphi}, \sigma^2 
    & \sim &
    \mathcal{N}(\boldsymbol{\varphi}, \sigma^2\mathbf{I}),
\end{eqnarray}
The GP model predicts \(f\) at unobserved points by updating the posterior mean, covariance, and variance using available data \(D_t = \{(x_i, y_i)\}_{i=1}^{t}\). These functions of the kernel hyperparameters \(\theta\), are computed as follows:
\begin{eqnarray}
    \mu_{\theta,t}(x) & = & \boldsymbol{k}_{\theta,t}(x)^\top \boldsymbol{C}_{\theta,t}^{-1} \boldsymbol{y}_t, \\
    k_{\theta,t}(x, x') & = & k_{\theta}(x, x') - \boldsymbol{k}_{\theta,t}(x)^\top \boldsymbol{C}_{\theta,t}^{-1} \boldsymbol{k}_{\theta,t}(x'), \\
    \sigma^2_{\theta,t}(x) & = & k_{\theta}(x, x) - \boldsymbol{k}_{\theta,t}(x)^\top \boldsymbol{C}_{\theta,t}^{-1} \boldsymbol{k}_{\theta,t}(x),
\end{eqnarray}
where the covariance matrix \(\boldsymbol{C}_{\theta,t}\) is defined as \(\boldsymbol{C}_{\theta,t} = \boldsymbol{K}_{\theta,t}(\boldsymbol{X}, \boldsymbol{X}) + \sigma^2\mathbf{I}_t\), incorporating both the structured knowledge from the kernel function, as parameterized by \(\theta\), and the observation noise with variance $\sigma^2$.
The marginal log-likelihood function, critical for iterative computing of the kernel hyperparameter \(\theta\) upon new data acquisition, is:
\begin{equation}
    \log p(\boldsymbol{y}_t | X_{1:t}, \theta) = -\frac{1}{2}(\boldsymbol{y}_t - \boldsymbol{m}_{\theta,t}(\boldsymbol{X}))^\top \boldsymbol{C}_{\theta,t}^{-1} (\boldsymbol{y}_t - \boldsymbol{m}_{\theta,t}(\boldsymbol{X})) %\nonumber \\
   -\frac{1}{2}\log |\boldsymbol{C}_{\theta,t}| - \frac{t}{2}\log(2\pi).
\end{equation}
%
%The computational complexity of evaluating this likelihood is \(\mathcal{O}(t^3)\), posing challenges for large datasets. 
The computational complexity of evaluating this likelihood is \(\mathcal{O}(t^3)\) posing challenges for large datasets.
To improve computational efficiency methods such as sparse Gaussian processes and high-dimensional Bayesian optimization are utilized as proposed by \citet{NIPS2005_sparse_GP_ghahramani, pmlr-nayebi2019_hd_BO} and \citet{Eriksson_NEURIPS2019_Turbo}.

%%%%%%%%%%%%%%%%%%%%%%%%%%%%%%%%%%%%%%%%%%%%%%%%%

\subsection{Bayesian optimization}
\label{subsec:BO}
Bayesian Optimization (BO) is a strategy for optimizing expensive-to-evaluate functions, often used in hyperparameter tuning of algorithms, as highlighted by \citet{Snoek_NIPS2012_hyperp1, Shahriari2016}. It employs Gaussian Process (GP)-based surrogate models to estimate the target function from available data. This estimate is iteratively updated thanks to each new data point \((x_t, y_t)\) at iteration \(t\). The essence of BO lies in its decision-making process for subsequent evaluations, leveraging acquisition functions on the GP posterior distribution to guide the search. BO aims to strike a balance between exploring new regions of the parameter space and exploiting areas with promising outcomes informed by prior observations. The surrogate model, pivotal to BO, incorporates a mean function \(\mu_{\theta}(\cdot)\) and a covariance function \(k_{\theta}(\cdot, \cdot)\), as detailed in section \ref{subsec:GP}. The iterative refinement of the model and subsequent decision-making are influenced by the hyperparameters \(\theta\). 
For simplicity, with little abuse of notation, subsequent mentions of \(\mu_{\theta}(\cdot)\) and \(k_{\theta}(\cdot, \cdot)\) will omit the subscript \(\theta\), though it remains part of the model's structure since $\theta$ is optimized as well.

% Acquisition function is important
The acquisition function is an important ingredient of most BO approaches since it estimates the potential scores of the next input values $x$ to be tested. The best value $x$ that optimizes this acquisition function is selected for the next evaluations of the objective function.
%
%% Acquisition functions
The GP-UCB (Gaussian Process Upper Confidence Bound), introduced by \citet{Srinivas2012_GPUCB}, is one example among various acquisition functions. It operates on the principle of maintaining a consistent confidence level, which is ideal for smoother underlying objective functions. The GP-UCB function is defined mathematically as:
\begin{equation}
    \alpha_{\text{GP-UCB}}^{(t)}(x) = \mu_{t-1}(x) + \sqrt{\beta_t} \sigma_{t-1}(x)
\end{equation}
where \( \beta_t \) represents a confidence parameter that is usually iteratively updated. % or, in the case of standard UCB, is fixed during the optimization procedure. 
The usual evolution of $\beta_t$ inspired by the work by \cite{srinivas_2010_ICML} is:
\begin{equation}
    \label{eq:betaevol-GPUCB}
\beta_t = 2\log\left(\frac{\pi^2}{3\delta} t^{2+d/2}\right)
\end{equation}
where $d$ is the dimension of the problem, and $\delta\in(0,1)$ has been fixed to 0.1 in our experiments.
When $\beta_t$ is fixed, the function is called UCB; Note that a common fixed $\beta$ value is 2.
%
% Expected Improvement
Expected Improvement (EI) is another widely used acquisition function that targets significant improvements over the best current observations \citep{Shahriari2016}. The EI function at iteration \( t \) is given by:
\begin{equation}
    \alpha_{\text{EI}}^{(t)}(x) = (\mu_{t-1}(x) - f(x^*_{t-1}) -\xi) \Phi(z) + \sigma_{t-1}(x) \phi(z)
\end{equation}
where \( f(x^*_{t-1}) \) is the best observed objective function value up to iteration \( t-1 \). The standardized variable \( z \) is defined as \( (\mu_{t-1}(x) - f(x^*_{t-1}) -\xi) / \sigma_{t-1}(x) \), where \(\sigma_{t-1}(x)\) is the standard deviation of the predicted distribution at point \(x\) at iteration \( t-1 \), and $\xi$ is a hyperparamter that permits us to balance between exploration and exploitation; larger values of $\xi$ favor exploration.
The terms \(\Phi(z)\) and \(\phi(z)\) represent the cumulative distribution function and the probability density function of the standard normal distribution, respectively.  This formulation of EI accounts for both the expected improvement over the current best observation and the uncertainty associated with the prediction.

%% Thompson Sampling (TS)
Thompson Sampling (TS) can be seen as a probabilistic acquisition function. It is an adaptive method for sequential decision-making that has been effectively integrated into BO frameworks with Gaussian Processes \citep{Kandasamy2018}. 
At each iteration \( t \), TS chooses \( x_t \) according to the following steps.
Let \( D_{t-1} = \{(x_i, y_i)\}_{i=1}^{t-1} \) represent the currently available query-observation pairs. The surrogate model $\tilde{f}_t$ of the objective function $f$ is defined as a Gaussian process. It is a random function distributed according to \( \mathcal{GP}(\mu_t, k_t) \), where the expectation $\mu_t$ and the covariance matrix $k_t$ are condtionned to the knowledge of $D_{t-1}$. 
Sampling a function that lives in an infinite dimensional space is not directly possible. Therefore, a usual method is to define a fine sampling grid over which values of the function can be sampled. These random variables all follow a Gaussian distribution that is easy to sample from. The corresponding continuous function can be reconstructed by interpolation for instance.
Then TS chooses the next test value $x_t$ that maximizes $\tilde{f}_t$. An advantage of TS is that it takes into account both a goal-orented information carried by $\mu_t$ and the current uncertainty carried by $k_t$ on the knowledge of the objective function $f$ from previous observations when sampling the surrogate model $\tilde{f}_t$. A known pitfall of TS is its computational cost in high dimensions.
%

%% Parallel BO
\subsection{Parallel Bayesian optimization}
\label{subsec:parallel_BO}
Parallel BO \citet{ginsbourger2008_qEI} extends the sequential optimization framework to accommodate the possibility of simultaneous evaluations of the objective function $f$, an adaptation for scenarios like HTE in chemical engineering. At each iteration \( t \), this approach permits the selection of a batch of points \( \mathbf{x}_{t,1:B} = \{x_{t,1}, x_{t,2}, ..., x_{t,B}\} \), rather than a single point as in traditional sequential BO. The objective function evaluations for this batch are then given by \( \mathbf{y}_{t,1:B} = f(\mathbf{x}_{t,1:B}) \), effectively transforming the optimization process into a multi-point search. 
This batch-wise approach is particularly beneficial in chemical engineering contexts, such as heterogeneous catalytic reactions, 
where multiple reactor blocks can be evaluated simultaneously, often with a shared constraint like reactant flow. 
Note that the adaptation of acquisition functions for batch selection in parallel BO is a key factor in addressing these unique challenges.

The GP-UCB-PE (Gaussian Process Upper Confidence Bound and Pure Exploration) algorithm, introduced by \citet{Contal2013}, is a two-phase process for batch selection in parallel BO using Gaussian Processes (GPs). The first phase utilizes the GP-UCB strategy for initial batch element selection, focusing on exploiting current knowledge (\(x_{t,0} = \arg\max_{x \in X} \mu_{t}(x) + \beta_t \sigma_{t}(x)\)). The subsequent phase shifts to pure exploration for the remaining batch elements, emphasizing regions of high uncertainty by maximizing updated posterior variance. %(\(x_{t,b} = \arg\max_{x \in X} \sigma_{t}^{(k)}(x)\)). 
This method effectively balances exploration and exploitation in batch evaluations and is particularly suitable to scenarios with high-uncertainty regions.

In the context of parallel BO, Thompson Sampling (TS) is adapted to select a batch of points simultaneously. \citet{Kandasamy2018} successfully incorporated TS to parallelize BO in their work. Unlike the sequential setting where a single function realization is sampled from the posterior, parallel TS involves sampling a batch of functions at each iteration \( t \):
\begin{equation}
    \mathbf{g}_t = \{g_{t,1}, g_{t,2}, ..., g_{t,B}\} \sim p(g|D_{t-1})
\end{equation}
where \( \mathbf{g}_t \) represents a batch of functions sampled from the posterior distribution conditioned on the data available up to iteration \( t\), a Gaussian process in the present context. Each function \( g_{t,b} \) in the batch is then optimized:
\begin{equation}
    x_{t,b} = \arg\max_x g_{t,b}(x)
\end{equation}
These points \( \{x_{t,1}, x_{t,2}, ..., x_{t,B}\} \) form the batch to be queried next. \citet{Kandasamy2018} demonstrate that this approach effectively translates the benefits of TS from the sequential to the parallel setting, allowing for efficient and robust exploration of the design space in a batch wise manner. TS permits a good trade-off between a goal-oriented sampling of the design-space and a good exploration of uncertain regions. 
This procedure is referred to as synchronous TS when one gathers first all batch results and then proceeds with
an update of \(D_{t}\). Therefore each step is informed by the accumulated data and the inherent probabilistic nature of GPs. 
TS also favors some diversity within a batch of samples.

%\pc{Here discussion on diversity within batch:}\\
In addition to previous methods, let us mention that \citet{Kathuria2016BatchedGP} contributed to the field of parallel BO by modeling the diversity of a batch using determinantal point processes (DPPs) that induce repulsive interactions between samples, thereby enhancing the efficiency of batch selection. Additionally, the work of \citet{pmlr-v70-wang17h} extends DPP-based acquisition functions to high-dimensional settings, proving instrumental in tackling high-dimensional batch problems. \citet{shah_NIPS2015_parallelEntropy} and \citet{Wang2016ParallelBG} further advanced this field with their development of novel algorithms for batch Bayesian optimization, showcasing the versatility of these methods in handling complex optimization scenarios. There is probably still room for improvement of the diversity of batch samples to optimize even further the exploration/exploitation trade-off.

%%%%%%%%%%%%%
\section{Process constrained parallel Bayesian optimization}
\label{sec:pcBO}

\subsection{Parallel BO in chemical engineering}
In chemical engineering problems, \citet{Gonzalez2023} introduced a decomposition approach for BO that enables efficient querying of the objective function per batch. 
This decomposition is led by a lower fidelity physics-based reference model. Their approach was tested extensively on numerical reactor case studies. 
\citet{Kondo2022} utilized BO for efficient parallel screening of the optimal working conditions for the synthesis of biaryl compounds, achieving high yields. Their method incorporates tailored acquisition functions and exemplifies the practical application of BO in enhancing sustainable manufacturing processes. 
Furthermore, \citet{Jose_Pablo2023} combines multi-fidelity 
modeling with asynchronous batch BO and showcases its 
efficiency in laboratory settings like battery design. There, experiments involve a multi-fidelity objective as well as data that differ in quality and time necessary to be generated. This approach significantly accelerates experimental 
processes by effectively utilizing lower-fidelity tests 
to predict higher-fidelity outcomes, optimizing resources and 
suggesting broader applications in fields with similar 
experimental challenges.

These developments underscore the importance and 
effectiveness of parallel BO in chemical engineering 
and various scientific disciplines. By enabling 
simultaneous evaluations and effectively incorporating process 
constraints, parallel BO has become a pivotal tool for optimizing experimental 
workflows, leading to significant advancements in automated 
experimentation and the design of experiments (DoE).

However, those application cases did not address optimization problems with {\em process constraints} as defined by \citet{Vellanki2017} and illustrated in Figure \ref{fig:hierarchy_scheme}. These experimental constraints are a critical aspect in certain chemical engineering scenarios. Traditional parallel BO methods often do not explicitly tackle such process-constrained optimization problems. Given this gap, our work shifts focus to specifically addressing process-constrained batch optimization challenges. This is particularly relevant to our setting in the REALCAT's Flowrence unit, where such constraints exist, see Figure \ref{fig:flowrence_scheme}.
%This is particularly relevant to our setting in the REALCAT's Flowrence unit, where such constraints play a vital role. T
The next section formally introduces the problem of process-constrained optimization in the context of parallel Bayesian Optimization, outlining the challenges and our approach to addressing them.

\subsection{Process-constrained batch optimization problem}
\label{subsec:pc-problem}

Process-constrained batch optimization (pc-BO) is a specific case within the hierarchical process-constrained framework introduced in section \ref{sec:Introduction}. This section focuses on pc-BO due to its direct applicability in many practical experimental setups, including those in REALCAT in the context of the MRS Flowrence system. 
The general form of the problem is given by the global optimization 
problem in Eq. \ref{eq:globOptimization}. In batch Bayesian optimization, this involves considering a sequence \(\{\boldsymbol{x}_{t,0}, \boldsymbol{x}_{t,1}, \dots, \boldsymbol{x}_{t,K-1}\}_{t=1}^T\), with each point \(\boldsymbol{x}_{t,k} = [\boldsymbol{x}^{uc}_{t,k}, \boldsymbol{x}^{c}_{t,k}]\) in the design space comprising unconstrained variables \(\boldsymbol{x}^{uc}_{t,k}\) and constrained variables \(\boldsymbol{x}^{c}_{t,k}\). 
Constrained variables \(\boldsymbol{x}^{c}_{t,k}\) remain constant for all \(k\) in a given batch, reflecting the process constraints; unconstrained variables \(\boldsymbol{x}^{uc}_{t,k}\) can be freely tuned. 
The design space is thus represented as \(X = X^c \times X^{uc}\), where \(X^c\) and \(X^{uc}\) denote the spaces of constrained and unconstrained parameters.
Eq. \ref{eq:globOptimization} becomes
\begin{equation}
    \boldsymbol{x}^* = \arg\max_{\boldsymbol{x} \in \mathcal{X}^{c}\times \mathcal{X}^{uc}} f(\boldsymbol{x}),
    \label{eq:pcOptimization}
    \end{equation}

Constraint BO methods, such as those proposed 
by \citet{Gelbart_2014_unkown_constraints}, 
 \citet{Gardner_2014_inequality_constraints}, 
and \citet{pmlr-v37-hernandez-lobatob15}, 
focus on different aspects of constrained optimization. 
Additionally, they are mainly designed for sequential or single-point optimization, 
making them less suited for the batch 
optimization challenges in chemical engineering and in HTE on MRS settings in particular.

To the best of our knowledge, \citet{Vellanki2017} were the first and only to 
introduce and formalize the concept of process-constrained 
batch Bayesian optimization introduced in Eq. \ref{eq:pcOptimization}. 
This problem involves scenarios where specific control variables, 
known as the constrained set, are costly or difficult to change 
and must remain fixed across a batch. Meanwhile, the remaining variables, 
the unconstrained set, are allowed to vary. This separation of the design space into two subsets of constraint and unconstrained parameters is a specific case of a hierarchical process-constrained framework with only two levels, with $\ell_0 = \mathcal{X}^c$ and $\ell_1 = \mathcal{X}^{uc}$, see Fig.~\ref{fig:hierarchy_scheme}. 
%I.e. $\mathcal{X}^c$ must be equal across all batch members as illustrated in Figure \ref{fig:hierarchy_scheme}. 

%Note that \citet{Vellanki2017} also used Gaussian Processes as the surrogate model in their approaches.

The first approach in \citet{Vellanki2017} is the pc-BO(basic) method that proposes a point for the first batch element at each iteration \(t\) by maximizing the GP-UCB-PE acquisition 
function over the complete design space, as detailed in 
section \ref{subsec:parallel_BO}. This is formulated as 
\(x_{t,0} = [x^{uc}_{t,0}, x^{c}_{t,0}] = \arg\max_{x \in \mathcal{X}} \alpha(x_{t,0} | D_{t-1})\). 
For subsequent elements, \(k = 1, \dots, K-1\), the algorithm fixes the constrained 
variable based on \(x^{c}_{t,0}\) and selects the unconstrained variables by 
maximizing their variance.
In contrast, the second method named pc-BO(nested) employs a nested optimization 
process with two separate GP models trained with separate datasets, $D_{t, I}$ and $D_{t,O}$ for the respective inner and outer data sets, to optimize both constrained and unconstrained variables in a batch. Initially, 
it identifies the optimal constrained variable \(x^c_t\) by 
maximizing \(\alpha\) over \(\mathcal{X}^c\) using 
the data set \(D_{t-1,O}\), formulated as 
\(x^c_t = \arg\max_{x^c \in \mathcal{X}^c} \alpha(x^c_t | D_{t-1,O})\). 
%
%\pc{Check for these notations : \(D_{t-1,O}\) and \(D_{t-1,I}\)?}
%
The algorithm then optimizes the unconstrained variables, starting 
with \(x^{uc}_{t,0}\), by maximizing \(\alpha^{uc}\) for the 
unconstrained variables for fixed \(x^c_t\). Subsequent batch 
elements are determined by maximizing the variance \(\sigma_{uc}\) 
to explore the unconstrained space, 
\(x^{uc}_{t,k} = \arg\max_{x_{uc} \in \mathcal{X}_{uc}} \sigma(x^{uc}_{t,k} | D_{t-1,I}, x^c_t, \{x_{uc}^{t,k'}\}_{k' < k})\). 
Both data sets \(D_{t-1,O}\) and \(D_{t-1,I}\) are updated 
with the outcomes of each batch's evaluations, allowing for an 
efficient exploration of both sets of variables.
Numerical tests by \citet{Vellanki2017} indicate that the pc-BO(nested) 
approach is slightly more efficient than pc-BO(basic). Additionally, 
pc-BO(nested) provides a convergence guarantee, a theoretical advantage 
that is not offered by the pc-BO(basic) method.
Empirical evaluations of \citet{Vellanki2017}'s pc-BO methods in our simulated specific experimental setup revealed that pc-BO(basic) frequently outperformed pc-BO(nested). This was established through a series of controlled numerical tests, aiming at a robust and reproducible evaluation framework. 

Based on these insights and the known efficiency and simplicity of Thompson Sampling in batch Bayesian optimization, we developed the pc-BO-TS method.
The development of pc-BO-TS is a direct response to the need for an efficient and effective method tailored to the unique challenges of process-constrained batch optimization. It extends the existing pc-BO framework and is specifically designed to enhance performance in practical HTE applications, particularly those involving complex multi-reactor systems.

\subsection{Process-Constrained Batch BO via Thompson Sampling}
\label{sec:optimizationAndRegret}

Thompson Sampling (TS), originally developed to address 
the multi-armed bandit problem is a 
Bayesian approach \citep{William_TS_1993}. TS is applicable in scenarios where sequential 
decision-making is needed under uncertainty. TS maintains a 
probabilistic model of the environment that is often represented by Gaussian processes. 

The fundamental principle of TS is to balance between exploration 
and exploitation. TS is conceptually simple, easy to implement, and also 
scales to a bigger number of parallel evaluations as described by \citet{Kandasamy2018}. Additionally, unlike deterministic methods that choose the 
next evaluation point based on the current best estimate,
TS employs a probabilistic approach. It randomly 
samples function instances from the model's posterior and selects the respective maximum of those drawn functions to determine the next evaluation batch. 

The pc-BO-TS algorithm~\ref{algo:pc-BO-TS} integrates TS to ensure a balanced and informed exploration of the design space. 
The iterations of the pc-BO-TS algorithm consist of several key steps. At iteration $t=0$, it initializes the set of evaluated points $X$ and their corresponding function values $\y$. Then the main loop of the algorithm runs until the maximum number of iterations $T$ is reached.
At each iteration, a batch of points $X_{\text{batch}}$ is formed, starting with the determination of the first batch member $\boldsymbol{x}_{t,0}$ by maximizing the Gaussian process Upper Confidence Bound (GP-UCB) over the domain $\mathcal{X}$:
\begin{equation}
\boldsymbol{x}_{t,0} = [\boldsymbol{x}^{uc}_{t,0}, \boldsymbol{x}^{c}_{t,0}] = \arg\max_{\boldsymbol{x} \in \mathcal{X}} \alpha_{UCB}(\boldsymbol{x}_{t,0} | D_{t-1}),
\end{equation}
where $\alpha_{UCB}$ is the UCB acquisition function, and $D_{t-1}$ is the current dataset of observations. The constrained component $\boldsymbol{x}_{t,0}^{c}$ of $\boldsymbol{x}_{t,0}$ is then fixed for the subsequent batch members, and $\boldsymbol{x}_{t,0}$ is added to $X_{\text{batch}}$.
For each remaining member of the batch (from 2 to $B$), the algorithm fixes the constrained parameters, creates a grid of the unconstrained ones, and employs Thompson Sampling to draw a sample from the Gaussian process function. Specifically, for each $k$ in $\{2, \ldots, B\}$, a set of function instances $\tilde{f}_{t,k}$ is sampled from the posterior of the GP conditioned on the current data $D_{t-1}$ and the fixed constrained variable $\x_{t,0}^{c}$:
\begin{equation}
\tilde{f}_{t,k} \sim \mathcal{GP}(\mu_{t-1}, \sigma_{t-1}^2 | D_{t-1}, \x^{c}_{t,0}),
\end{equation}
Samples $\x_{t,k}$ of next batches will inherit values constrained by $\x^{c}_{t,0}$:
\begin{equation}
    \forall k\in \{2, \ldots, B\}, \: \x^{c}_{t,k} = \x_{t,0}^{c}.
\end{equation}
Then the point that maximizes sampled function $\tilde{f}_{t,k}$ within the unconstrained space $\mathcal{X}^{uc}$ is identified and added as the next batch member. It is such that:
\begin{equation}
\x^{uc}_{t,k} = \arg\max_{\boldsymbol{x}^{uc} \in \mathcal{X}^{uc}} \tilde{f}_{t,k}([\x^{uc}, \x_{t,k}^c]),
\end{equation}
The objective function $f$ is then evaluated at all the resulting batch points $\x_{t,k}=[\x^{uc}_{t,k}, \x^{c}_{t,k}]$. The dataset is updated with these new observations. The Gaussian process model is refined with the new batch of data. This process ensures a meticulous search of the design space, balancing exploration and exploitation, and adhering to process constraints. The algorithm concludes once the maximum number of iterations is reached, returning the set of evaluated points $X$ and their corresponding function values $\boldsymbol{y}$.

\begin{algorithm}%[h!]
    \caption{Process-Constrained Batch Bayesian Optimization with Thompson Sampling (pc-BO-TS) \label{algo:pc-BO-TS}}
    \begin{algorithmic}[1]
    \Require $f$: Objective function, $\mathcal{X}$: Design space, $B$: Batch size, $T$: Maximum number of iterations
    \Ensure $X$: Evaluated points, $\boldsymbol{y}$: Function values
    \State Initialize evaluated points $X$ and function values $\boldsymbol{y}$
    \State Set iteration counter $t \leftarrow 0$
    \While{$t < T$}
        \State Initialize batch $X_{\text{batch}} \leftarrow \emptyset$
        \State Determine the first batch member $\boldsymbol{x}_{t,0} \leftarrow \arg\max_{\boldsymbol{x} \in \mathcal{X}} \alpha(\boldsymbol{x}_{t,0} | D)$
        \State Extract and fix constrained component $\boldsymbol{x}_{c}^{t,0}$
        \State Add $\boldsymbol{x}_{t,0}$ to the batch $X_{\text{batch}} \leftarrow X_{\text{batch}} \cup \{\boldsymbol{x}_{t,0}\}$
        \For{$k = 1 \text{ to } B-1$}
            \State Sample a function instance $\tilde{f}_{t,k} \sim \mathcal{GP}(\mu_{t-1}, \sigma_{t-1}^2 | D_{t-1}, \boldsymbol{x}^{c}_{t,0})$
.            \State Determine $k$-th batch member $\boldsymbol{x}^{uc}_{t,k} \leftarrow \arg\max_{\boldsymbol{x}^{uc} \in \mathcal{X}^{uc}} \tilde{f}_{t,k}(\boldsymbol{x}^{uc})$
            \State Add $\boldsymbol{x}_{t,k} \leftarrow [\boldsymbol{x}^{uc}_{t,k}, \boldsymbol{x}^{c}_{t,0}]$ to the batch $X_{\text{batch}} \leftarrow X_{\text{batch}} \cup \{\boldsymbol{x}_{t,k}\}$
        \EndFor
        \State Evaluate objective function at batch points $\boldsymbol{y}_{\text{batch}} \leftarrow f(X_{\text{batch}})$
        \State Update dataset with new observations $D_{t} \leftarrow D_{t-1} \cup \{(X_{\text{batch}}, \boldsymbol{y}_{\text{batch}})\}$
        \State Update Gaussian process model with $X_{\text{batch}}$ and $\boldsymbol{y}_{\text{batch}}$
        \State Increment iteration counter $t \leftarrow t + 1$
    \EndWhile
    \State Return evaluated points and function values $X, \boldsymbol{y}$
    \end{algorithmic}
\end{algorithm}
%%% END ALGO 1

\subsection{Hierarchical process-constrained batch optimization via Thompson sampling (hpc-BO-TS)}
\label{subsec:H-pc-BO-TS}

\begin{algorithm}
    \caption{Tree-Structured Hierarchical Process-Constrained Bayesian Optimization (HPC-BO)}
    \label{algo:tree_hpc_bo}
    \begin{algorithmic}[1]
    \Require $f, \mathcal{X}, T, N, \{K_\ell\}_{\ell=0}^N$
    \Ensure $(\x^*, \y^*)$
    \State $\mu_0 = 0_d$  
    \State $\mathbf{x}_{0,0} = \text{Uniform}(\mathcal{X})$ \Comment{Random uniform initialization}
    \State $D_0 \leftarrow \emptyset$ \Comment{Initial dataset}
    \State $t \leftarrow 0$
    \While{$t \leq T$}
        \State $X_t^0 \leftarrow \{\x_{t,0}\}$
        \For{$\ell = 1$ to $N-1$}
            \State $X_t^\ell \leftarrow \{\x_{t,0}\}$ \Comment{Initialize level with UCB point}
            \State $\z = \x_{t,0}$
            \For{$k = 1$ to $K_\ell-1$} \Comment{Vary parameters for UCB point}
                \State Sample $\tilde{f}_{t,k} \sim \mathcal{GP}(\mu_t, \sigma_t^2 | D_{t-1})$
                \State $\x_{t,k}^{0:\ell-1} \leftarrow \z^{0:\ell-1}$
                \State $\x_{t,k}^{\ell:N} \leftarrow \argmax_{\x^{\ell:N}\in \prod_{i=\ell}^N {\cal X}_i} \tilde{f}_{t,k}(\z^{0:\ell-1},\x^{\ell:N})$
                \State $X_t^\ell \leftarrow X_t^\ell \cup \{\x_{t,k}\}$
            \EndFor
            \ForAll{$\z \in X_t^{\ell-1}\backslash \{\x_{t,0}\}$} \Comment{Vary parameters for other parent nodes}
                \For{$k = 0$ to $K_\ell-1$}
                    \State Sample $\tilde{f}_{t,k} \sim \mathcal{GP}(\mu_t, \sigma_t^2 | D_{t-1})$
                    \State $\x_{t,k}^{0:\ell-1} \leftarrow \z^{0:\ell-1}$
                    \State $\x_{t,k}^{\ell:N} \leftarrow \argmax_{\x^{\ell:N}\in \prod_{i=\ell}^N {\cal X}_i} \tilde{f}_{t,k}(\z^{0:\ell-1},\x^{\ell:N})$
                    \State $X_t^\ell \leftarrow X_t^\ell \cup \{\x_{t,k}\}$
                \EndFor
            \EndFor
        \EndFor
        \State $X_t \leftarrow X_t^{N-1}$, $Y_t \leftarrow f(X_t)$ \Comment{Evaluations at t}
        \State $D_t \leftarrow D_t \cup \{(X_t, Y_t)$\} \Comment{Update dataset}
        \State $\mathbf{x}_{t+1,0} = \arg\max \alpha_{UCB}(\mathbf{x} | D_{t})$
        \State $t \leftarrow t + 1$
    \EndWhile 
    \State \Return $(\x^*, \y^*) \leftarrow \argmax_{(\x,\y)\in D_T} \y(\x)$ \Comment{Final solution}
    \end{algorithmic}
\end{algorithm}
    
This section generalizes the pc-BO-TS approach for a hierarchical multi-reactor system with several levels, as illustrated in Fig.~\ref{fig:hierarchy_scheme}. It introduces the hierarchical process-constrained batch optimization via Thompson sampling (hpc-BO-TS).
%
% %### Overview of the Algorithm 2
Algorithm~\ref{algo:tree_hpc_bo} describes hpc-BO-TS. It extends the above pc-BO-TS to a hierarchy of more than 2 levels using a recursive tree-structured exploration strategy. The proposed strategy integrates UCB and Thompson sampling steps to balance the exploration of the search space with the exploitation of acquired information, efficiently navigating through the parameter hierarchy. 

% %# Initialization
In the absence of prior information, the first point \( \mathbf{x}_{0,0} \) is uniformly sampled from the entire design space \( \mathcal{X} \). This point is the basis for the hierarchical exploration at the first level. The prior on the surrogate objective function remains a Gaussian Process (GP) model, with mean function $\mu_{0}=0_{d}$ and covariance given by the initial kernel $k_0$.

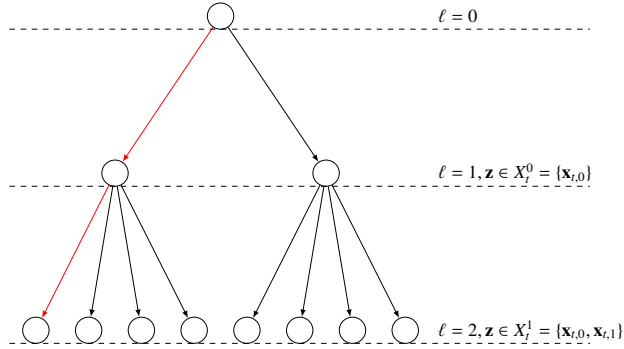
\begin{figure}
    \centering    
\resizebox{.5\textwidth}{!}{
\begin{tikzpicture}[grow=down, sloped]
\tikzset{
  treenode/.style={align=center, draw=black, circle, minimum size=.5cm, inner sep=.5pt},
  level 1/.style={level distance=3cm, sibling distance=4cm},
  level 2/.style={level distance=3cm, sibling distance=1cm},
  edge from parent/.style={draw=black, -latex}
}
  \node[treenode] (root) {}
    child {node[treenode] {} edge from parent[red]
      child {node[treenode] {} edge from parent[red]}
      child {node[treenode, draw=black] {} edge from parent[black]}
      child {node[treenode] {} edge from parent[black]}
      child {node[treenode] {} edge from parent[black]}
    }
    child {node[treenode] {}
      child {node[treenode] {}}
      child {node[treenode] {}}
      child {node[treenode] {}}
      child {node[treenode] {}}
    };
    
  \draw[dashed] (-4,-0.25) -- (7,-0.25);
  \draw[dashed] (-4,-3.25) -- (7,-3.25);
  \draw[dashed] (-4,-6.25) -- (7,-6.25);

  \node[draw=none, right] at (4,0.0) {$\ell=0$};
  \node[draw=none, right] at (4,-3.0) {$\ell=1, \z \in X_{t}^{0}=\{\x_{t,0}\}$};
  \node[draw=none, right] at (4,-6) {$\ell=2, \z \in X_{t}^{1}=\{\x_{t,0}, \x_{t,1}\}$};
\end{tikzpicture}}
    \caption{Tree structure of the dataset $X_t$ used by the hierarchical process-constrained algorithm where $\x_{t,0}$ results from UCB-based optimization (in red), determined at line 27 in \ref{algo:tree_hpc_bo}.}
    \label{fig:hpc_tree}
\end{figure}

% %# General recursive approach explanation
Given a budget of $T$ iterations, Algorithm \ref{algo:tree_hpc_bo} follows a recursive strategy illustrated by the tree structure of Figure~\ref{fig:hpc_tree}. 
The general idea of the recursion is that at each level $\ell$ of the hierarchy,  a new batch of $K_\ell$ samples is drawn. However, these new samples inherit a subset of (constrained) parameters from their parent sample, denoted by $\z$ in the algorithm. Therefore new samples $\x_{t,k}$ in a batch are such that $\x_{t,k}^{0:\ell-1}$ are inherited and kept fixed while the free components $\x_{t,k}^{\ell:N}$ are optimized.
Thompson sampling is used in this respect, and the data set $D_t$ is updated along iterations. Once the budget has been fully used, a final estimate of the searched optimum of the objective function $f$ is built. 

Figure~\ref{fig:hpc_tree} illustrates the recursive nature of algorithm~\ref{algo:tree_hpc_bo}. It corresponds to a specific case with \(N=3\) levels and batch sizes $(K_{\ell})_{0\leq \ell\leq 2} = (1, 2, 4)$. Each parent node \(\z\in X_t^{\ell-1}\) has children that form a batch of size $K_\ell$.
% %# Variables definition
The initial point determined via UCB $\x_{t,0}$ is the initial parent node $\z$: it is dealt apart (lines 9-15) through the full hierarchy, see the red branches. Other parent nodes $\z$ at a given level $\ell$ will determine the constrained parameters from the previous levels $0:\ell-1$ via $\x_{t,k}^{0:\ell-1} = \z^{0:\ell-1}$ (lines 16-22).
%
% %# Introducing different functioning for loops
This is the reason why Algorithm \ref{algo:tree_hpc_bo} features two different but very similar {\em for} loops. 
The first {\em for} loop (lines 10-15) ensures that the parameters $\x_{t,0}$ determined by UCB inform the selection of unconstrained parameters by using $\x_{t,0}$ as the parent node down to the final leaf, see the red branch in Figure \ref{fig:hpc_tree}. 
The second {\em for} loop (lines 16-22) considers all other general parent nodes apart from the UCB-determined point $\x_{t,0}$. This loop deals with the determination of all explored samples besides the one determined via UCB. Both loops follow the same logic of determining unconstrained parameters via TS.

% %# Tree expansion via TS + UCB for next iteration
Thompson Sampling is used to sample surrogate objective functions \(\tilde{f}_{t,k}\) in a similar manner in both loops. At level $\ell$, for all samples within a batch \(k = 0\) or $1$ (to keep the UCB sample apart) to $K_{\ell}-1$, the selection of free parameters \(\mathbf{x}_{t,k}^{\ell:N}\) maximizes \(\tilde{f}_{t,k}\) under the constraint that \(\mathbf{x}_{t,k}^{1:\ell-1} = \z^{0:\ell-1}\) from the parent node. This batch $\{\x_{t,k} = (\x_{t,k}^{0:\ell-1}, \x_{t,k}^{\ell:N}), \forall k\}$ contributes to the set \(X_t^\ell\) that will constrain the next level $\ell+1$. %The second for loop follows the same principle but for all other parent nodes.
%
% %# Data set update
Once the full hierarchy has been explored down to the leaves of the tree, the set of points $X_t$ is ready for evaluation of the objective function $f$. The resulting $X_t$ and $Y_t$ are added to the dataset $D_t$ for the next iteration $t+1$. 
%
% %# UCB lead exploration
From iteration \( t=1 \) to the total budget $T$, the UCB strategy selects the next initial point \( \mathbf{x}_{t+1,0} \) based on previous data $D_{t-1}$ for each new batch. The GP model is updated at each iteration with new observations, which enhances its predictive accuracy and refinement for subsequent iterations.

% %# Exception from the UCB lead exploration at t=0
Note that except at $t=0$, the initial point $\mathbf{x}_{t,0}$ is set by maximizing the UCB based on previous observations, serving as a potentially good predictor of the optimal point. Algorithm~\ref{algo:tree_hpc_bo} preserves this point as a leaf of the tree, meaning that $f(\x_{t,0})$ will be evaluated. This corresponds to the bottom left leaf with red connections in Fig.~\ref{fig:hpc_tree} and justifies the separation between lines 10-15 and 16-22 in Algorithm~\ref{algo:tree_hpc_bo}, even though they are very similar.

% %# Final estimate  
Eventually, an estimate of the optimal set of parameters \( \x^* \) can be determined. One simple strategy is to pick the one in $D_T$ that yields the highest empirically observed value of the objective function \( \y^* \). Another approach is to use a new surrogate function based on $D_T$ to estimate $\x^*$.

As a proof of concept, section~\ref{subsec:syntheticTestCase_hpc} presents an experimental application of the method on a known function.

%%%%%%%%%%% SYNTHETIC CASe STUDIES / EXPERIMENTS
\section{Synthetic Case Studies}
\label{sec:syntheticTestCase}

Synthetic test cases are essential for a comprehensive evaluation of the optimization methods introduced in this study. Section~\ref{subsec:2d_Synthetic_Test_case} deals with the 2-dimensional setting where visualization remains easy and illustrative. Section~\ref{subsec:multidimensional_synthetic_test_cases} will deal with more complex objective functions up to  dimension 6.

\subsection{2D Synthetic Test Case Studies}
\label{subsec:2d_Synthetic_Test_case}
Here we introduce two-dimensional synthetic test cases based on Gaussian Mixture models (GMMs). They are designed within a controlled two-dimensional design space, \(D = \{x \in \mathbb{R}^2: -3 \leq x_i \leq 3 \text{ for } i \in \{1,2\}\}\). Such generative models provide a standardized environment to assess the performance of the algorithms across various scenarios that mimic the complexity encountered in real-world objective functions, in particular the presence of several modes or extrema.

\subsubsection{Description of the synthetic test case generation}
\label{subsubsec:description_synthetic_test_case}
The synthetic objective functions are generated using Gaussian Mixture Models (GMMs), which are capable of approximating the yield objective functions of heterogeneously catalyzed reactions. A GMM is defined as a weighted sum of Gaussian distributions:
\begin{equation}
    p(\mathbf{x}) = \sum_{k=1}^{K} \pi_k \mathcal{N}(\mathbf{x} | \mathbf{\mu}_k, \mathbf{\Sigma}_k)
\end{equation}
where \( p(\mathbf{x}) \) is the probability density function at point \( \mathbf{x} \) and \( K \) is the number of Gaussian components; the \( \pi_k \) are the mixture weights and \( \mathcal{N}(\mathbf{x} | \mathbf{\mu}_k, \mathbf{\Sigma}_k) \) denotes the individual Gaussian distribution with mean vector \( \mathbf{\mu}_k \) and covariance matrix \( \mathbf{\Sigma}_k \).

Utilizing a sequential algorithm, one objective function is generated for each case, with the case number corresponding to the number of means in the GMMs. 
% Means
The means $\mu_k$ have been randomly and uniformly sampled over $[-3,3]^2$ in all cases.
% Covariances
% Case 1 to 3
For cases 1 to 3, covariance matrices were randomly sampled with coefficients uniformly distributed in the range [0.7, 1.3], ensuring distinct local optima by maintaining adequate spacing between the means, i.e., modes. 
% Case 4
In contrast, case 4 features a smoother single-maximum landscape with significant overlap of the Gaussian components, achieved by selecting the covariance matrix coefficients from a wider range of [1.5, 2.0].
% The details of the algorithm, including its pseudo-code, are provided in the Appendix.

\begin{figure}%[H]
    \centering
    % First row
    \begin{subfigure}{0.45\textwidth}
        \centering
        \includegraphics[width=\linewidth]{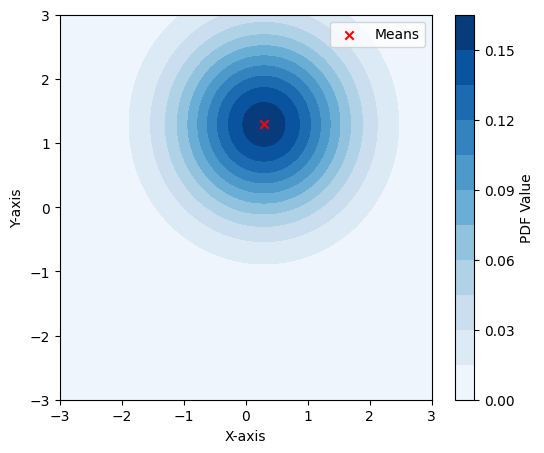}
        \caption{Gaussian mixture model case 1}
        \label{fig:gmm1}
    \end{subfigure}%
    \hfill
    \begin{subfigure}{0.45\textwidth}
        \centering
        \includegraphics[width=\linewidth]{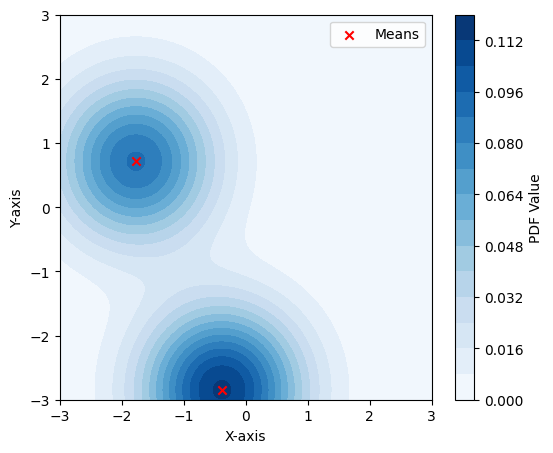}
        \caption{Gaussian mixture model case 2}
        \label{fig:gmm2}
    \end{subfigure}
    
    % Second row
    \begin{subfigure}{0.45\textwidth}
        \centering
        \includegraphics[width=\linewidth]{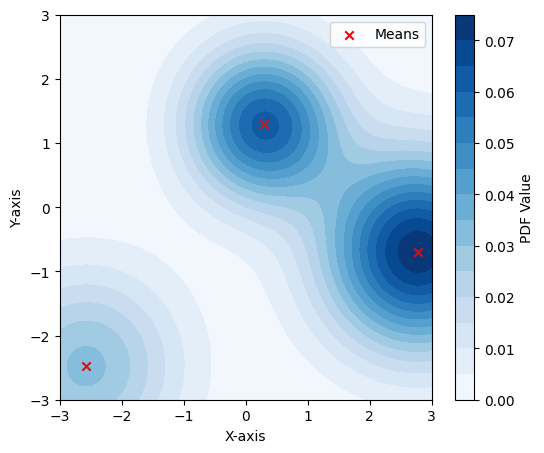}
        \caption{Gaussian mixture model case 3}
        \label{fig:gmm3}
    \end{subfigure}%
    \hfill
    \begin{subfigure}{0.45\textwidth}
        \centering
        \includegraphics[width=\linewidth]{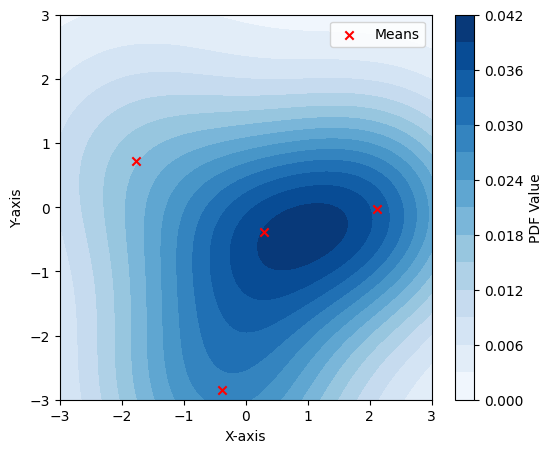}
        \caption{Gaussian mixture model case 4}
        \label{fig:gmm4}
    \end{subfigure}
    
    \caption{Randomly generated Gaussian mixture model based objective functions.}
    \label{fig:gmm_cases_instances}
\end{figure}

Figure \ref{fig:gmm_cases_instances} displays the generated two-dimensional GMM objective functions for all cases. The red cross locates the centers $\mu_k$ of each Gaussian component of a mixture. 
%The indices of cases correspond to the respective orders of generated GMMs. 
These synthetic GMM-based objective functions serve as the basis for 
a series of optimization tests conducted in dimension 2, which permits pedagogical illustration as well.

\subsubsection{Experimental setup and performance metrics}
\label{subsubsec:experimental_setup_evaluation}

In this study, various process-constrained Batch Bayesian Optimization (pc-BO) methods are rigorously evaluated through a series of synthetic two-dimensional test cases. These cases are crafted to reflect the complexities and challenges encountered in real-world optimization problems, particularly in the realm of chemical engineering and experimental settings.

The constraint in our experimental setup is applied to the first parameter \(x_{c} = x_1\), keeping it fixed, while the second parameter \(x_{\text{uc}} = x_2\) remains unconstrained. The methods under evaluation encompass a wide range of optimization strategies, including pc-BO(nested)-GPUCB and UCB, pc-BO(basic)-GPUCB and UCB, pc-TS-UCB, pc-TS-EI, sequential Bayesian optimization, and random sampling.
For all methods, a batch size of $B=4$ was chosen for all test runs on GMM's.
%% UCB vs GP-UCB
When using GP-UCB, the usual evolution of $\beta_t$ is governed by \eqref{eq:betaevol-GPUCB}.
For UCB, in subsequent experiments, the value of $\beta$ is fixed to 2.

%%%
The synthetic test cases are generated using Gaussian Mixture Models (GMMs), where for each test case one GMM is generated. The GMMs exhibit varying means and covariances, yet their optima and covariance matrices ranges are maintained within a similar range for each case to ensure consistency and comparability across different runs and GMMs.

To thoroughly assess the performance of the optimization methods, each method was executed 10 times on every objective function, using the same set of random initial points for all methods within their respective classes. For pc-BO approaches, the first point is sampled uniformly randomly from the design space, while the remaining $K-1$ points are sampled from the constrained subspace, ensuring consistency within this class with $K$ being the batch size. In contrast, in GP-UCB-PE, all $K$ points are sampled uniformly randomly from the design space due to its different functioning. The very first point is the same as in the pc-BO methods. This approach ensures a robust evaluation, capturing the algorithms' behavior across various conditions and highlighting their strengths and weaknesses.

%% log normalized regret
The performance of the optimization methods is quantitatively evaluated using the normalized regret of eq.~\eqref{eq:normalized-risk} in section~\ref{subsec:seq-optim}. The normalized regret ranges from 0 to 1, the smaller the better. In practice, to favor a finer interpretation of the results, the log-normalized regret is used to put emphasis on the detailed performance of a method when the normalized regret gets close to the optimal 0 value. 
All the functions used in the following experiments have been translated to take values between 0 and their maximum. Therefore the log-normalized regret can be easily read as a relative precision. A log-normalized regret of -1 corresponds to a precision of 10\%, -2 to 1\%, -3 to 0.1\%, etc.
Since each method was run 10 times with varying initializations we chose to visualize the median of the log normalized regret to demonstrate their respective performance.
The analysis of the distribution of log-normalized regrets at the last iteration via kernel density estimates (KDE) for a set of initializations for each method will also permit us to study the robustness of the methods. Distinct modes that are far apart will indicate that a method got stuck in local optima. Those metrics, i.e. median log normalized regret and KDEs will be used throughout all experiments in this work.
%%%%%%%%
\subsubsection{Results and findings}
\label{subsubsec:synthetic_results_and_findings}

\begin{figure}%[h!]
    \centering
    % First row
    \begin{subfigure}{0.65\textwidth}
        \centering
        \includegraphics[width=\linewidth]{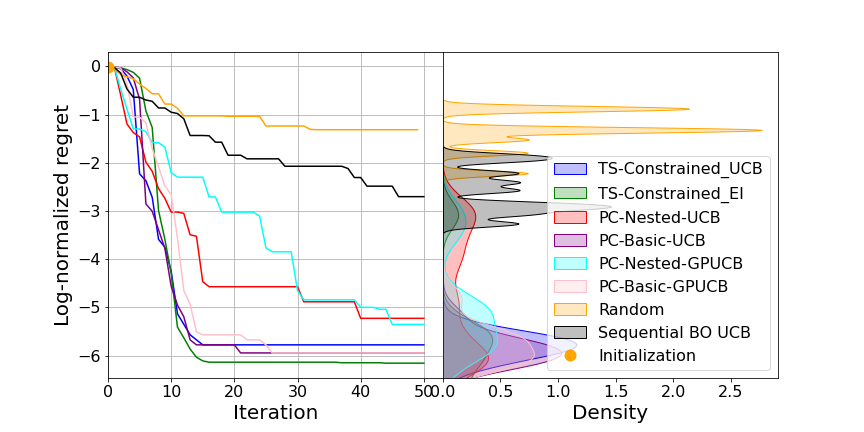}
        \caption{Gaussian mixture model case 1}
        \label{fig:gmm_case_1}
    \end{subfigure}%
    \hspace{-5pt}
    \begin{subfigure}{0.65\textwidth}
        \centering
        \includegraphics[width=\linewidth]{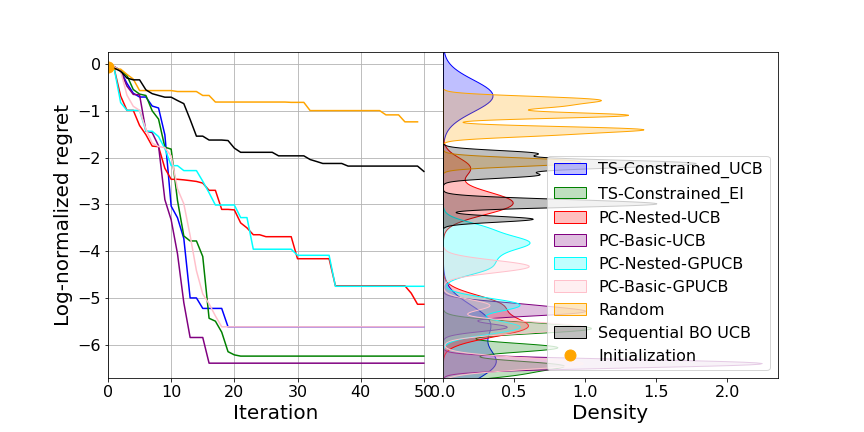}
        \caption{Gaussian mixture model case 2}
        \label{fig:gmm_case_2}
    \end{subfigure}
    % Second row
    \begin{subfigure}{0.65\textwidth}
        \centering
        \includegraphics[width=\linewidth]{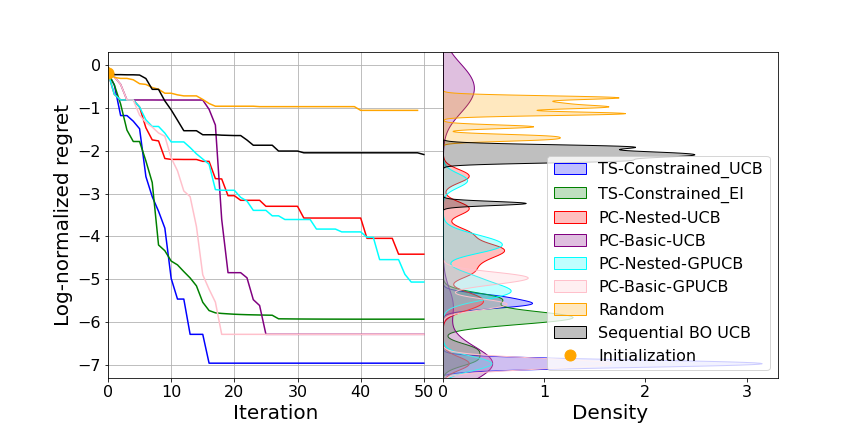}
        \caption{Gaussian mixture model case 3}
        \label{fig:gmm_case_3}
    \end{subfigure}%
    \hspace{-5pt}
    \begin{subfigure}{0.65\textwidth}
        \centering
        \includegraphics[width=\linewidth]{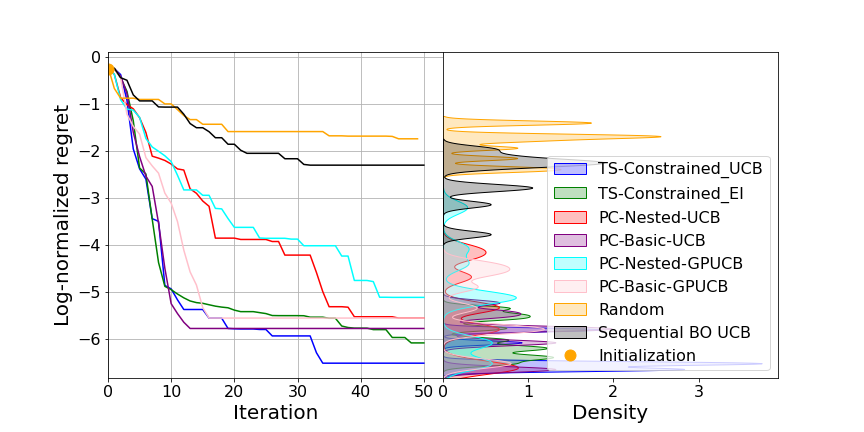}
        \caption{Gaussian mixture model case 4}
        \label{fig:gmm_case_4}
    \end{subfigure}
    
    \caption{Evaluation of optimization methods across different Gaussian mixture model cases. Presented are the median log normalized regrets on the right and the kernel density estimates of the final iteration regrets on the left for all methods. Both figures share the same y-axis.}
    \label{fig:gmm_cases}
\end{figure}

Figure~\ref{fig:gmm_cases} illustrates the behavior of a set of methods in the scenarios described above.
The asymptotic regret reached by a method after a large number of iterations illustrates its accuracy with respect to the true optimum.
The convergence speed of the optimization methods provides insight into 
their efficiency and effectiveness in reaching optimal solutions. 
The distribution of their regret reflects their robustness linked to their sensitivity to initial conditions.
In applications, a finite accuracy is often sufficient, i.e. 1 or 0.1\% for instance. The budget for experiments, i.e. iterations, is often a practical limitation. Robustness to initial conditions is important since only one run of experiments may be implemented. Therefore a good trade-off between accuracy, speed and robustness is targeted.
%

%Unimodal
%Initial convergence
In the unimodal case 1 in Figure \ref{fig:gmm_cases}(a) TS-Constrained\_UCB shows the most rapid initial convergence, with a log normalized regret of -2 reached at around iteration 5. All pc-BO and pc-TS methods show rapid initial convergence with all methods reaching a regret level of -2 within 10 iterations. 
%Asymptotic behavior
All methods show good asymptotic behavior with all methods eventually exceeding an optimum accuracy of -5 regret. With TS-constrained\_EI reaching the highest accuracy with a regret of -6. 
% Robustness discussion
The KDEs reveal distinct behaviors among the different optimization methods in terms of the distribution of their final regret, indicating varying levels of robustness to initial starting conditions. While TS-Constrained\_UCB and PC-Basic-UCB exhibit high density around the final -6 regret level, suggesting robustness, PC-Nested-UCB displays a wider spread distribution ranging from -2 to -6, making it more susceptible to initial conditions. Note that the KDE y-axes being the same as the log normalized regret occasionally cuts off the full distribution, as is the case with TS-Constrained\_UCB, which has some density beyond the -6 regret level. PC-Basic-UCB also demonstrates high robustness with a high density at the desirable -6 regret level. For all metrics, Random and single sequential BO perform worst which is due to either their respective strategy or that only one observation is conducted per iteration. 
% Theorize in terms of exploration exploitation trade off why each methods behaves how.
The PC-Basic approaches perform comparably with the pc-BO-TS approaches in terms of robustness, asymptotic behavior, and initial convergence.  This shows that the in principle more exploratory strategy of the pc-BO basic methods are as robust in this case as the more goal-oriented pc-BO-TS approaches. 

%%Duomodal Case 2
%Initial Convergence
In the bimodal case 2 in Figure \ref{fig:gmm_cases}(b) all batch methods initially converge fast. They
reach a regret of -2 within the first 10 iterations.
A regret of -3 is reached within 9-12 iterations by all pc-BO(basic) and pc-TS methods.
Note that a log normalized regret of $-3$ already means a relative precision of $0.1\%$ with respect to the optimum value $f(x^*)$. Such precision, combined with a low variance, can be sufficient in many applications.
%Asymptotic behavior
Between iteration 15-20 the pc-BO(basic) and pc-TS approaches come close to their final regret and start to plateau between -5.5 to -6.5. The pc-BO(nested) methods converge significantly slower and reach a higher regret around -5.
% Robustness discussion
The KDEs for TS-constrained\_EI and PC-Basic-UCB showcase high density around a regret level of -6. While the pc-BO(nested) methods show clearly a wider distribution of final regrets with PC-Nested-UCB having two distinct modes around a regret of -5 and one around -2.5. This indicates higher sensitivity towards initial starting points. However, the TS-Constrained-UCB KDE shows clearly that this method got stuck in the local optimum by reaching several times a regret between 0 to -1. This could be caused by the less exploratory strategy employed in general by the pc-TS methods. 

%Trimodal Case 3
In Case 3 in Figure \ref{fig:gmm_cases}(c), which introduces additional complexity, the pc-TS methods initially converge the fastest. TS-Constraiend\_UCB and TS-Constrained\_EI reach a very good level of -4 regret within the first 10 iterations. The next method showing moderately fast initial convergence is PC-Basic-GPUCB reaching a regret level of -4 at iteration 15. 
%Asymptotic behavior
As in the previous cases, the methods reaching or exceeding the lowest regret around -6 are the pc-BO(basic) and pc-TS methods. TS-Constrained\_UCB shows the best final regret with -7. 
% Robustness discussion
The KDEs show distinct patterns for the pc-BO and pc-TS methods. Generally, the pc-BO(nested) methods show a wide spread of regrets indicating less robustness, while the pc-TS methods show little sensitivity in this case. One specifically distinct pattern is the PC-Basic-UCB KDE which shows that the PC-Basic-UCB method gets several times stuck in a local optimum with a regret not exceeding -2. This shows that even though the pc-BO approaches generally have a more exploratory strategy, they are sensitive to initial starting points and can be prone to getting stuck at a local optimum. 

%% Case 4
In the unimodal Case 4 in Figure \ref{fig:gmm_cases}(d), where the PDFs are not distinct from each other, the pc-TS methods emerge as the top performers, 
underscoring their effectiveness across varying test scenarios. 
Across all cases, Random and sequential Bayesian optimization lag behind, showcasing their limited utility in these specific optimization challenges. 
In summary, the analysis of convergence speed and 
robustness across different synthetic test cases reveals distinct 
trends and performance characteristics of the evaluated optimization 
methods. 

The pc-TS and pc-BO(basic) methods consistently demonstrate rapid convergence and robust performance, even in complex scenarios. 
Nevertheless, note in case 3 PC-Basic-UCB exhibits slower convergence compared to the other cases. Therefore both Thompson sampling approaches and in general the pc-BO(basic) maximum variance approaches in combination with a strategy balancing exploration versus exploitation can be used to successfully solve process-constrained optimization problems.
% Revision 2 Comment 1
The use of either EI or UCB shows favorable performance within all GMM test cases. This demonstrates the robustness of the pc-TS method with respect to the choice of the acquisition function.
In contrast, the pc-BO(nested) methods consistently perform asymptotically worse and are less robust. random 
and sequential Bayesian optimization methods lag behind the most, 
highlighting their limitations for these particular optimization challenges.

\subsection{Multi-dimensional synthetic test cases}
\label{subsec:multidimensional_synthetic_test_cases}
This section will study the influence of the number of constrained parameters. 
The increase in dimensionality exacerbates the issues of sparsity and volume in the search space, distorts our intuitive understanding of distance and proximity, and complicates the optimization landscape with local optima. 
For Bayesian optimization, these challenges increase the computational needs and make the optimization of the acquisition function more difficult.

The number of constrained parameters can be seen as a proxy parameter indicating how parallel the optimization approach is. If all parameters are 
unconstrained, the optimization problem can be tackled by a fully parallel Bayesian optimization approach. If all parameters are constrained it can only be tackled by a single query Bayesian optimization approach. 
In order to assess the performance of the pc-BO algorithms as well as the effect of the number of constrained parameters, performance is evaluated by the log normalized regret\footnote{All the functions have been chosen to be positive over the considered domain, with values ranging from 0 to the maximum value.}. 
Furthermore, all experiments are run for an identical set of 10 random initial points. By default, the number of constrained parameters is 
always $n_c = d/2$. 
A batch size of $B$ = 4 is chosen in this section for all parallel methods. \ref{subsec:multidimensional_synthetic_test_cases}.
% 
%From now the performance is evaluated by the log regret instead of the normalized log regret form, i.e. the logarithm of Eq. \ref{eq:cumulativeRegret}. 

\subsubsection{Levy and Hartmann 6D results and discussion}
\label{subsubsec:levy_hartmann_results}

% \subsection{Results and Discussion on Multi-Dimensional Test Cases}

The selection of the Levy 6D and Hartmann 6D functions as benchmarks 
for optimization stems from their challenging multi-modal landscapes, which 
are widely recognized for testing the efficacy of optimization 
algorithms in higher dimensions. The inherent complexity of 
these functions, marked by multiple local optima, poses a 
significant test to the balance between exploration and exploitation strategies in Process-Constrained 
Bayesian Optimization (pc-BO) methods.

\begin{figure}%[h]
    \centering
    % First row
    \begin{subfigure}{\textwidth}
        \centering
        \includegraphics[width=\linewidth]{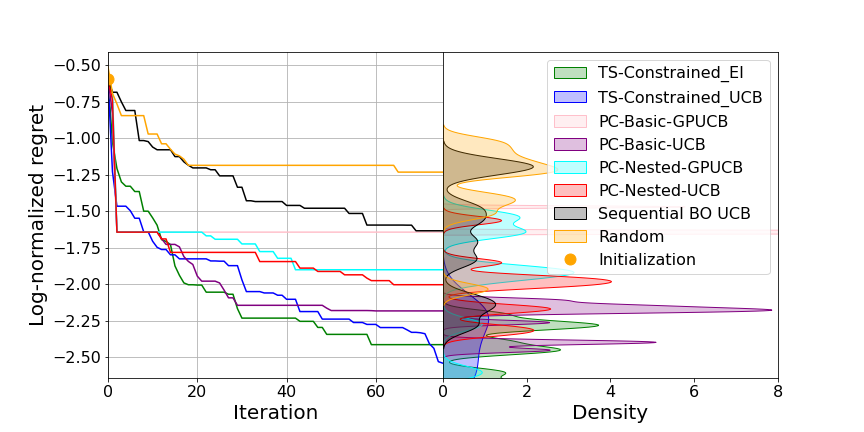}
        \caption{Levy 6D.}
        \label{fig:levy}
    \end{subfigure}

    % \hfill
    \begin{subfigure}{\textwidth}
        \centering    
        \includegraphics[width=\linewidth]{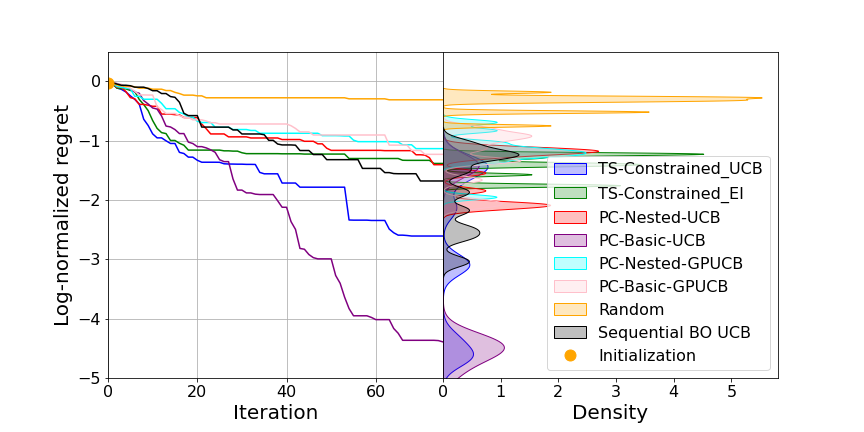}
        \caption{Hartmann 6D.}
        \label{fig:hartmann}
    \end{subfigure}
    \label{fig:levyhartmann}
    \caption{Process-constrained method performances on Levy and Hartmann 6D test functions with 3 constrained dimensions among 6: (left) evolution of the median of the log-normalized regret over iterations; (right) distribution of the last iteration optima estimated by each method. Both figures share the same y-axis.}
\end{figure}

% \paragraph{Analysis of the Styblinski-Tang Function}
% Initial and final median observed regret analyzed
Figure~\ref{fig:levy} shows the performance of various optimization strategies on the Levy 6D function. All pc-BO methods initially converge quickly towards the global optimum, but by iteration 15, their behaviors begin to diverge. PC-Basic-GPUCB levels off at a log regret of about -1.6, while PC-Nested-GP-UCB and PC-Nested-UCB continue to improve, ending with final log regrets of -1.9 and -2.0, respectively. Remarkably, by iteration 17, TS-Constrained\_EI achieves a median log regret of -2, reaching 1\% median regret accuracy in just 17 iterations. PC-Basic-UCB reaches this regret level by iteration 25, followed by TS-Constrained\_UCB at iteration 30. TS-Constrained\_EI, TS-Constrained\_UCB, and PC-Basic-UCB effectively converge towards the optimum, achieving final log normalized regrets of about -2.5, -2.4, and -2.2, respectively. Sequential BO, although slower initially, reaches the same regret level of -1.6 as PC-Basic-GPUCB.

% Analyze the robustness in terms of stability (how often does method reach the same result) and variance 
Figure~\ref{fig:levy} also presents the KDEs of each method's final log normalized regret at iteration T=75. This helps us discuss each method's robustness and likelihood of getting stuck in local optima. The KDEs confirm the final log normalized median regret trends. TS-Constrained\_UCB, TS-Constrained\_EI, and PC-Basic-UCB show a high concentration of probability around a low regret of -2 to -2.5. Particularly, TS-Constrained\_UCB shows a smooth and even density in this range, often exceeding a regret of -2.5. In contrast, PC-Basic-UCB appears more likely to get stuck in a local optimum, as indicated by its sharp peaks within the -2 to -2.5 regret range. TS-Constrained\_EI, with less pronounced peaks, seems less prone to getting stuck at a local optimum.

All other methods show densities around higher log normalized regrets, confirming the earlier observed trends in normalized regret.

% \paragraph{Analysis of the Hartmann 6D Function}
Figure \ref{fig:hartmann} shows the optimization results for the Hartmann 6D function. TS-Constrained\_UCB shows moderately fast convergence, reaching a regret of -1 by about iteration 15. However, PC-Basic-UCB takes the lead by iteration 27, consistently improving and ultimately reaching a final log regret below -4. All other methods achieve lower median final regrets, with the single-query UCB approach coming in third. This result suggests that larger batch sizes may be necessary in higher dimensions to adequately cover the exponentially increasing search space volume, highlighting the complex relationship between batch size and dimensionality in pc-BO methods.

The KDEs of the final regret values in Figure \ref{fig:hartmann} reveal distinct patterns. While the density distributions confirm the superior median results of TS\_Constrained\_UCB and PC-Basic-UCB, they also show high variance for these methods. TS-Constrained\_UCB's final regrets are spread between -1 and -5, exhibiting several local peaks. In contrast, PC-Basic-UCB displays fewer peaks: a minor one around -1 and a more prominent one around -4.5. This suggests that the exploratory approach of PC-Basic-UCB reaches the global optimum more frequently.

% \paragraph{Summary}
The multi-dimensional test cases demonstrate that pc-TS and pc-BO(basic) methods can 
achieve robust performance against complex, multi-modal benchmarks. 
However, the findings also highlight the critical impact of the acquisition function 
and batch size in high-dimensional optimization.
TS-Constrained methods for instance exhibit rapid convergence and efficiency, as demonstrated by TS-Constrained\_EI in the Levy 6D function. These methods balance exploration and exploitation effectively, outperforming PC-Basic and PC-Nested approaches in early iterations for both objective functions. While PC-Basic-UCB tends to get trapped in local optima, indicated by distinct modes in KDE plots in the Levy 6D case, TS-Constrained methods show broader regret distributions, suggesting a more adaptive search strategy. The less exploratory approach and thus more optimum-oriented approach of TS-Constrained methods enhances sample efficiency by reducing evaluations in less promising regions. However, it may limit thorough exploration, potentially impacting long-term optimization potential. This contrast highlights the trade-off between the rapid convergence benefits of optimum-oriented strategies and the exploratory depth provided by continuous exploration of high uncertainty regions, as seen in PC-Basic and PC-Nested methods.

Future work may focus on enhancing the exploration capabilities of 
pc-BO methods and optimizing batch sizes to better navigate the challenges presented 
by high-dimensional spaces.

\subsubsection{Effect of the number of constrained parameters}
\label{subsubsec:effect_of_constrained_parameters}

\begin{figure}%[h!]
    \centering
    \begin{subfigure}[b]{0.7\textwidth}
        \centering
        \includegraphics[width=\textwidth]{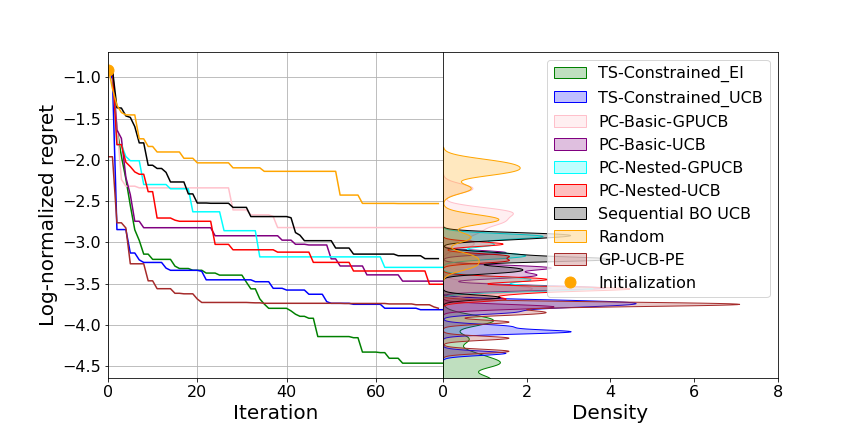}
        \caption{One constrained variable.}
        \label{fig:one_var}
    \end{subfigure}
    
    \begin{subfigure}[b]{0.7\textwidth}
        \centering
        \includegraphics[width=\textwidth]{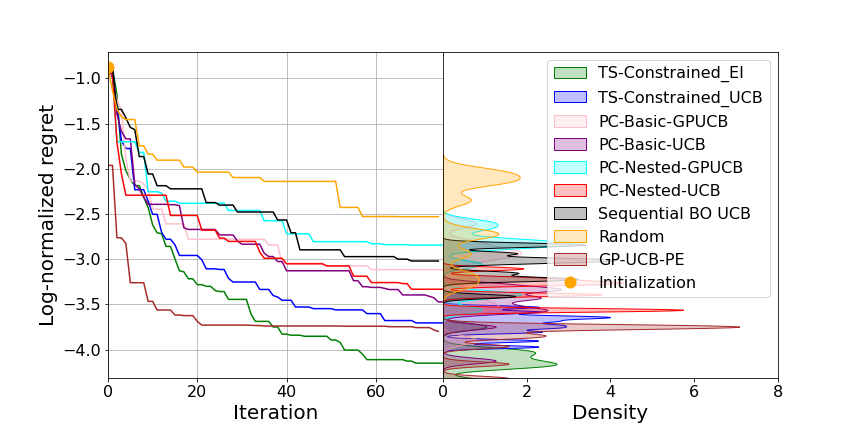}
        \caption{Two constrained variables.}
        \label{fig:two_var}
    \end{subfigure}
    
    \begin{subfigure}[b]{0.7\textwidth}
        \centering
        \includegraphics[width=\textwidth]{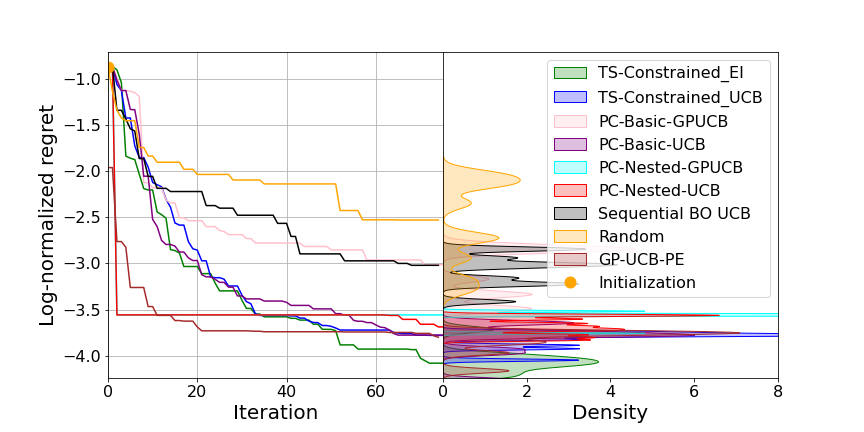}
        \caption{Three constrained variables.}
        \label{fig:three_var}
    \end{subfigure}
    
    \caption{Median log normalized regret convergence and final iteration regret distributions visualized for process-constrained BO methods and parallel BO using GP-UCB-PE on the Rosenbrock 4D function, across three scenarios. Both figures share the same y-axis.}
    \label{fig:rosenbrock_tests}
\end{figure}
%%%%

% Motivation of experiment
The impact of the number of constrained parameters on Bayesian optimization performance was analyzed using the Rosenbrock 4D function, known for its challenging landscape. This function allows for varying the number of constraints within the computational limits of the Thompson sampling method. The study compares the performance of pc-BO methods under different constraint scenarios with the fully parallel BO approach, GP-UCB-PE by \citet{Contal2013}, introduced in \ref{subsec:parallel_BO}, to evaluate the cost implications of constraints relative to a fully parallel approach.
%
% Key findings stated at the beginning
The key finding of this section is that increasing the number of constrained parameters reduces the advantage of using parallel methods like the TS-Constrained methods. This point is critical for understanding the behavior of different methods under varying constraint conditions.

% observations case 1 variable constrained
Figure~\ref{fig:rosenbrock_tests}(a) illustrates the performance with just one constrained parameter. Initially, all methods steadily converge towards the optimum, with the free GP-UCB-PE showing the fastest initial convergence, achieving a log normalized regret of -3.5 by iteration 10. The rapid convergence of TS-Constrained\_EI and TS-Constrained\_UCB, similar to GP-UCB-PE, indicates that, with few constraints, these methods can exploit the search space efficiently. This advantage will decrease as the number of constraints increases. Note that TS-Constrained\_EI reaches the best asymptotic accuracy.

% observations case 2 variable constrained
Figure~\ref{fig:rosenbrock_tests}(b) shows that all process-constrained methods exhibit a slower but steady convergence pattern when the number of constrained parameters increases to two. Despite this, TS-Constrained\_EI and TS-Constrained\_UCB maintain a lead through the iterations, still demonstrating an effective navigation in the constrained parameter space, although with slower initial progress compared to the single-constraint scenario. The corresponding distribution of the log-normalized regret remains concentrated around its median which illustrates a good robustness.

% observations case 3 variable constrained
Figure~\ref{fig:rosenbrock_tests}(c) with three constrained parameters shows an even slower convergence of all process-constrained methods. This is expected since the higher the constraints, the more difficult the optimization process. The spread in performance increases corresponding to a weakened robustness as well. TS-Constrained\_EI achieves the best final regret, around -4.0, closely followed by TS-Constrained\_UCB and PC-Basic\_UCB.%, and PC-Nested-GPUCB. 
Note that PC-BO-Nested-UCB seems to exhibit very good behavior with high initial accuracy. As soon as iteration 3, PC-BO-Nested-UCB proposes querying at $[0,0,0,0]$ where the objective function is only 3 below the true optimum that is 10827 at $x^*=[1,1,1,1]$, see eq. \eqref{eq:rosenbrock_app_4d}. Thus this apparent good performance is misleading since the method did not find the actual optimum  $x^*$.  The regret is therefore very low by chance, while the method proposes a spurious location of the optimum. Indeed, the central location $[0,0,0,0]$ is proposed by default when the optimization of the acquisition function fails within PC-BO-Nested-UCB, due to the strong importance of constraints here. This issue with the AF optimization step is a known challenge in BO \citep{Wilson2018MaximizingAF, Shahriari2016}.

% Generall trends observed in terms of asymptotic accuracy, robustness and budget
General trends are observed across all 3 cases. 
% Asymptotic behavior
Despite initial slower convergence under higher constraints, goal-oriented methods like TS-Constrained\_EI and TS-Constrained\_UCB achieve asymptotically smaller final regrets. These methods effectively balance exploration and exploitation, ultimately leading to lower regrets even when the number of constraints increases. 
% Budget efficiency 
Within a finite budget, which is most often the case in practice, note that TS-Constrained\_EI and TS-Constrained\_UCB both exhibit very good behavior. They reach a normalized regret as small as $10^-3$ within less than 20 iterations. A determination of the searched optimum with a relative precision of $0.1\%$ is feasible within a very reasonable budget. We emphasize that a budget in iterations translates into a budget in resources in practice.
% Robustness
In terms of robustness, the narrow distribution of regrets indicates that methods like TS-Constrained\_EI consistently demonstrate a stable behavior across varying constrained scenarios, maintaining lower variance and smoother convergence patterns compared to other methods. This robustness is crucial in practical settings where achieving consistent performance is essential.

% Summarizing and reiterating the main observation: Higher number of constraints deteriate methods perfomance
As expected, the results indicate that increasing the number of constrained parameters reduces the exploration potential within the search space. This puts the emphasis on the critical role of robust methods that can balance exploration and exploitation effectively. The free GP-UCB-PE 
exhibits the expected best performance, especially in terms of initial convergence. It is used as a baseline. Throughout all scenarios, the pc-BO-TS approaches strike a better balance compared to all non GP-UCB-PE methods. This is illustrated by faster convergence and increased robustness with near-optimal regrets, which remains true regardless of the chosen acquisition function.

%---------------------- HPC EXPERIMENTAL SETTING  ----------------
\subsection{A multi-level problem for hierarchical process constrained optimization}
\label{subsec:syntheticTestCase_hpc}

This section illustrates the good behavior of the hierarchical hpc-BO-TS proposed in section~\ref{subsec:H-pc-BO-TS} in a simple synthetic case, the Rosenbrock3D objective function, that serves as a proof of concept for an application over 3 levels.

\subsubsection{Experimental setting}
\label{subsec:expe_setting_hpc}

The hpc-BO-TS approach is applied to the Rosenbrock function, defined over a three-dimensional space. The function is known for its narrow, curved valley structure, which poses a challenge for optimization algorithms due to its multiple local optima. The hierarchy of constraints is aligned with the 3 dimensions. 
%% Rosenbrock 3D
The objective function denoted by \(f: \mathbb{R}^3 \rightarrow \mathbb{R}\), is (the opposite of) a version of the Rosenbrock 3D function adapted to fit between 0 and its maximum value, see \ref{subsec:synthetic-functions}. 

%% Comparison with GP-UCB-PE
The unconstrained Gaussian Process Upper Confidence Bound with Pure Exploration (GP-UCB-PE) method \cite{Contal2013} will be used as a baseline for comparison. Based on GPs and UCB, it seems to be a good reference for a fair comparison. A common Matérn kernel parameterized by \(\nu = 2.5\) is used for both methods. The GP's hyperparameters, specifically the kernel parameters, are updated at each iteration from \(t\) to \(t+1\) by maximizing the log-likelihood of the observed data. Initially, all mean functions \(\mu\) are set to zero. All used UCB acquisition functions use a fixed $\beta = 2.0$ hyperparameter.  
GP-UCB-PE is also expected to potentially yield a better performance than hpb-BO-TS which has limited exploration power due to the presence of constraints. 

The hierarchical set of constraints is the same as in section~\ref{subsec:H-pc-BO-TS} with batch sizes $(K_{\ell})_{0\leq \ell\leq 2} = (1, 2, 4)$ and a total batch size of $8$, the number of leaves of the corresponding tree, see fig.~\ref{fig:hpc_tree}. 
The GP-UCB-PE strategy \cite{Contal2013} uses batches of size \(K_{\text{GP-UCB-PE}} = 8\).
The number of iterations is fixed to \(T = 75\), for  
\(N = 15\) independent trials to explore the robustness 
and reliability of the results. The initial points are the same for both methods. The performance of each strategy is 
quantified using the log-normalized regret metric,

\subsubsection{Results from hpc-BO-TS}
\label{subsec:expe_discussion}

%% FIGURE HPC BO
\begin{figure}
    \centering
    \includegraphics[width=.8\textwidth]{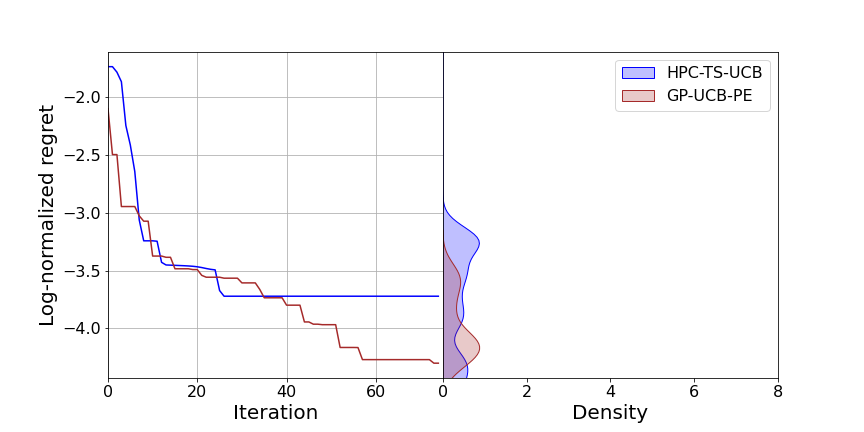}
    \caption{Performance comparison of optimization strategies over 75 iterations. The median log regret and variance are shown for hpc-BO-TS, GP-UCB-PE, sequential BO, and a Random Strategy. Both figures share the same y-axis.
    }
    \label{fig:results_hpc}
\end{figure}
%% END FIGURE

%%%%
Figure \ref{fig:results_hpc} shows a comparison between the results obtained by the proposed hierarchical process-constrained Bayesian optimization with Thompson sampling (hpc-BO-TS) and GP-UCB-PE.

As expected, GP-UCB-PE performs best since it benefits from the highest number of degrees of freedom.
However, note that hpc-BO-TS yields similar results after only a few iterations (about 6 or 7) with a good level of regret reached of -3. Then, GP-UCB-PE again shows better asymptotic behavior beyond 50 iterations with a final regret exceeding -4, which is expected from its larger latitude of exploration. Furthermore, the KDE indicates an increased variance for the HPC-TS-UCB approach which indicates that the constraints do not permit as robust convergence towards the optimum as GP-UCB-PE. 
For many applications in practice, the non-asymptotic short-term behavior is crucial. The desired precision on the estimate of the optimum is also often about 1\%, rarely smaller. 
Turning to the variance that is indicative of the method's sensitivity to initial conditions, it is only slightly higher for the hpc-BO-TS approach than for GP-UCP-PE. This may again stem from the presence of constraints. %This may lead to inconsistency in performance across different runs of the optimization process.
Therefore, this numerical illustration shows that the proposed hpc-BO-TS may provide very interesting performance in context of constrained systems such as multi-reactor systems.

%\clearpage
\subsection{Summary of synthetic case studies}
\label{subsec:summary_synthetic_cases}

Gaussian Mixture Models (GMMs) were used to create complex synthetic objective functions, simulating yield functions in heterogeneous catalysis.

Various pc-BO methods were assessed using these synthetic cases. Performance was assessed through normalized regret, a measure of relative optimization efficiency.
%    \item \textit{\ref{subsubsec:synthetic_results_and_findings} Results and Findings:} 
PC-Basic-GPUCB and TS-Constrained\_EI consistently perform best compared to other methods in terms of rapid convergence and robust performance across diverse synthetic test cases. PC-Basic-UCB also shows strong performance in robustness, while Random and Sequential Bayesian Optimization methods exhibit limitations.
%\end{itemize}

%\textbf{\ref{subsec:multidimensional_synthetic_test_cases} 
To test multi-dimensional synthetic cases, we used
%\begin{itemize}
%    \item 
\textit{Levy 6D and Hartmann 6D functions}\footnote{All these functions have been symmetrized and translated if necessary to make them range from 0 to their maximum value}. These functions test the efficacy of optimization algorithms in higher dimensions. The findings underscore the importance of the acquisition function and batch size in high-dimensional optimization. Notably, TS-Constrained\_UCB and PC-Basic-UCB converge fastest in these test cases. 
%    \item \textit{\ref{subsubsec:effect_of_constrained_parameters} Constrained Parameters Number Effect:} 
The performance of Bayesian optimization methods varies with the number of constrained parameters as shown in the \textit{rosenbrock 4D} case. As expected, an increase of the number of constrained parameters deteriorates the efficiency of process-constrained methods.%, highlighting the importance of exploration. 
TS-Constrained\_EI and TS-Constrained\_UCB perform best across all cases of numbers of constrained parameters, highlighting that the Thompson Sampling approach properly explores the design space. 
%\end{itemize}

%\textbf{Overall Findings and Implications:}
The experiments above presented the strengths and weaknesses of various Bayesian optimization methods within contexts of process constraints. Notably, the pc-BO-TS and the general hierarchical hpc-BO-TS approaches demonstrate good qualities in terms of rapid convergence and robustness. These good properties make them appropriate candidates for the optimization of multi-reactor systems where quick optimum convergence is needed.

%%%%%%%%%%%%%%%%%%%%%%%%%%%%%%%%%%%%%%%%%%%%
\section{Realistic case study}
\label{sec:realistic_case_study}

% \subsection{Introduction to REALCAT, HTE, and Validation Grid}
\subsection{The REALCAT platform and the concept of Digital Catalysis}
REALCAT is a one-of-a-kind state-of-the-art HTE platform dedicated to the development of catalysts \citep{RealCatWeb}. It integrates advanced
robotic technology and analytical devices for the design and parallel testing 
of catalysts under varied conditions, streamlining the 
catalyst development process. This efficiency aligns 
with the principles of Digital Catalysis, which 
focuses on accelerating research through improved data reproducibility 
and efficient data management.

As part of the French "équipements d'excellence" (EQUIPEX), REALCAT's 
goal is to enhance the efficiency of catalysts development, 
incorporating signal processing methods into its workflow. Its HTE 
capabilities, crucial for simultaneous catalyst and condition testing, 
exemplify the platform's role in advanced catalyst 
research. This sets the stage for our numerical tests of the 
process-constrained batch BO methods introduced before using 
experimental data to construct a surrogate model based on 
realistic experimental data.

\subsection{Experimental set-up}
\label{subsec:experimental_setup}

\begin{figure}%[H]
    \centering
    \includegraphics[width=0.63\textwidth]{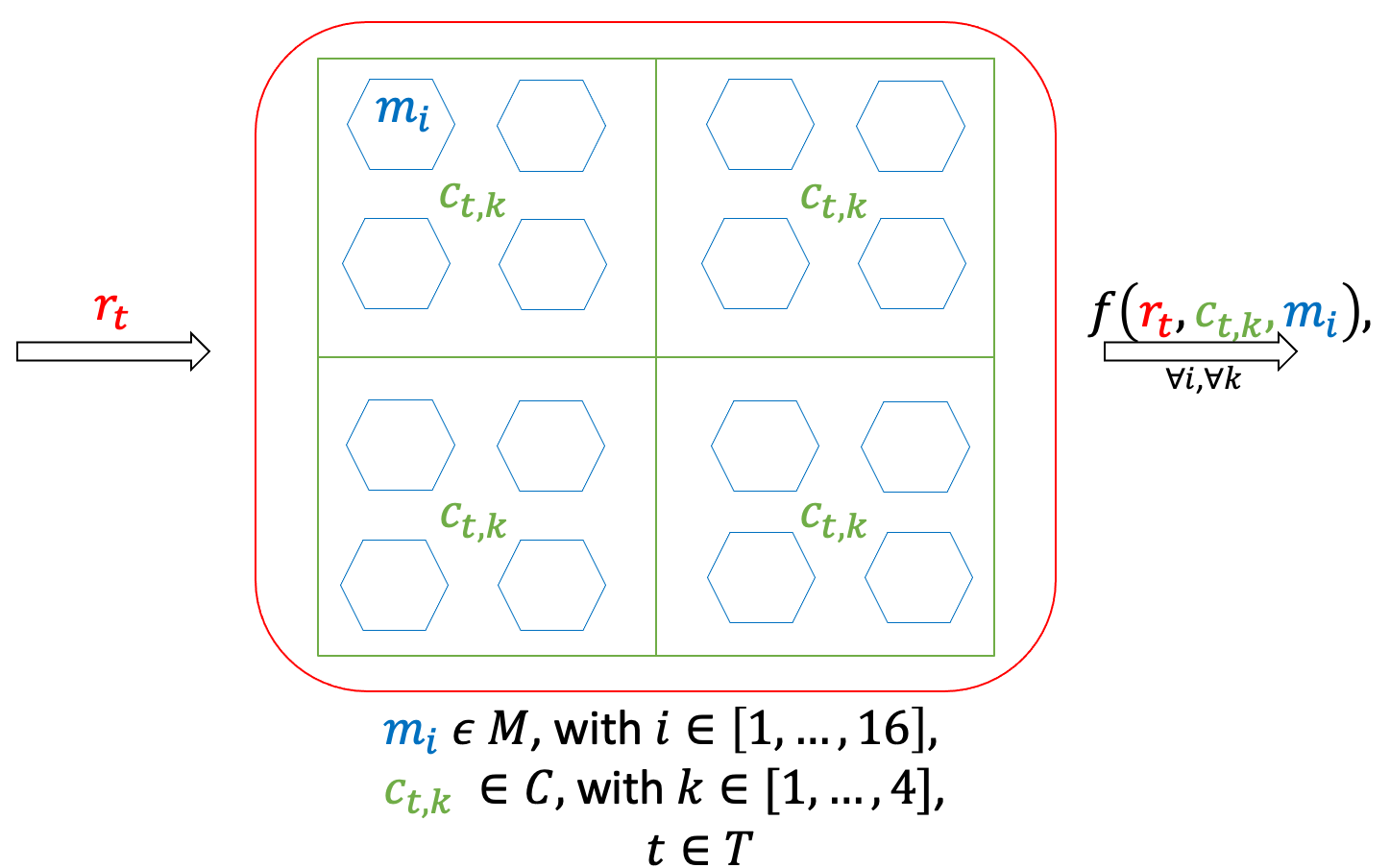}
    \caption{Operational scheme of the Flowrence unit displaying the different levels of degree of freedom of each parameter according to the hpc scheme in \ref{fig:hierarchy_scheme}.}
    \label{fig:flowrence_scheme}
\end{figure}

Within the context of process-constrained batch Bayesian optimization, the empirical validation of our method necessitates acquiring experimental data. For that, we utilized a Flowrence unit from AVANTIUM, detailed schematically in Figure \ref{fig:flowrence_scheme}. This multi-reactor device comprises sixteen fixed-bed tubular reactors $i$ loaded with different masses \( m_i \) of catalyst and arrayed into four blocks $k$, each offering independent thermal regulation \( c_{t,k} \), where \( t \) indexes the experimental optimization step. The flow of reactants \( r_t \) is uniformly distributed across all reactors. The true yield function  $f(r_t, m_i, c_{t,k})$ is thus a product of the interplay between the reactant flow rate, the mass of the catalyst within each reactor, and the specific temperature conditions of each block—parameters that are critical for optimizing catalytic activity.

Our model reaction, the oxidative dehydrogenation of propane (ODHP) to propylene, was chosen based on the industrial relevance it could have and the thermodynamic advantages it offers over propane dehydrogenation \citep{Gambo2021}. The catalyst selection was informed by Zhou et al., who identified isolated boron zeolite-based catalysts as high performers for ODHP due to their robust activity and selectivity \citep{Zhou2021}. 

The experimental design for the grid validation sequence was driven by the goal to construct a realistic surrogate of the ODHP yield function. We selected our experimental grid to explore the region within the design space where \cite{Zhou2021} indicated the presence of an optimal activity zone. Consequently, the validation grid spanned flow rates from \(16 \, \text{ml/min}\) to \(46 \, \text{ml/min}\) (atmospheric pressure, 20$^oC$, all flow rates in this work were calibrated under those conditions), temperatures between \(542^\circ C\) and \(590^\circ C\) and a mass configuration of 0 mg, 50 mg, 100 mg and 150 mg across all blocks. The masses were kept the same during the grid validation due to the high time cost associated with mass alterations within the Flowrence unit. These parameters were considered to enclose the optimal operating conditions for the ODHP reaction, as per prior empirical insights. The unconverted reactants and products of the reactions were analyzed online by chromatography.

\subsection{The surrogate objective function}
\label{subsec:surrogate_objective_function}

For the purpose of a systematic validation of the proposed approach with respect to some ground truth, a surrogate objective function is built. This is carried out by learning a GP model from a set of experiments corresponding to a sufficiently refined grid of parameters. The learned expectation $\mu_{GP}(x)$ of the Gaussian process is then used as the surrogate objective function $f_{\text{ref}}(x)=\mu_{GP}(x)$. This function can be considered as a realistic representation of the real problem. 

The model encapsulates the experimental observations, mapping the multidimensional relationship between the input parameters and the propylene yield. It was trained using the data from the validation grid (the validation grid dataset and the code used can be found at \href{https://github.com/Markusmu93/J-CompChE-HPCBoTS}{J-CompChE-HPCBoTS}). Experiments were run for a fixed mass of 150 mg, known to be an optimal choice, because it is difficult to change this mass on the fly in the Flowrence multi-reactor system. Indeed, the only parameters that can be changed on the fly on the Flowrence unit are the reactant flow and the block temperatures. Another reason to keep the mass constant was to make the problem two-dimensional for sake of pedagogical illustration.

% Revision 2 minor comment 4
The surrogate objective function \( f_{\text{ref}}(x) \) is the mean function \(\mu_{GP}(x)\) of the Gaussian Process model trained on a comprehensive set of experimental data. This function serves as a realistic approximation of the true yield function \( f \), capturing the relationship between the input parameters and the propylene yield. The surrogate model \( f_{\text{ref}}(x) \) thus enables to simulate the ODHP reaction within a controlled empirical framework. In summary, the dataset collected over a fine grid feeds the surrogate model \( f_{\text{ref}}(x) \) which, in turn, provides the controlled empirical framework for validation of our pc-BO-TS method.

\subsection{Problem Formulation}
The Flowrence unit at the REALCAT, designed for HTE in heterogeneous catalysis, operates as an MRS. While this system inherently presents a three-level HPC  problem, reactant flow uniformity across reactors, individual temperature control per block, and reactor-specific catalyst mass loading (as detailed in Figure \ref{fig:flowrence_scheme} and Subsection \ref{subsec:experimental_setup})—our experimental approach simplifies it into a two-level process-constrained batch optimization problem. This simplification is due to practical constraints: we fix the catalyst mass to avoid time-consuming reactor unloading, loading, and potential inaccuracies in recalibration with each iteration. Consequently, we only train the surrogate objective function with 150 mg catalyst mass which corresponds to maximum yield based on our grid analysis, see section \ref{subsec:experimental_setup}. Thus, the design space \(\mathcal{X}\) of the ODHP reaction on the Flowrence unit is effectively divided into constrained flow rate space \(\mathcal{X}^c\) and unconstrained temperature space \(\mathcal{X}^{uc}\), forming a Cartesian product \(\mathcal{X} = \mathcal{X}^c \times \mathcal{X}^{uc}\). This configuration aligns with our focus on process-constrained batch optimization as defined in Section \ref{subsec:pc-problem}, with the Flowrence unit's architecture dictating a batch size of four due to its four independently temperature-regulated blocks.

The surrogate model \( f_{\text{GP}}(x) \), a Gaussian Process trained on the high-yield mass data, serves as an approximation of the propylene yield function \( f(x) \). Our aim is to optimize \( f_{\text{GP}}(x) \), subject to the constraints of the Flowrence unit, to maximize the yield. The optimization objective is defined as:

\begin{equation}
\boldsymbol{x}^* = \arg\max_{\substack{x^c \in \mathcal{X}^c, \\ \{x^{uc}_{(k)}\}_{k=1}^B \in \mathcal{X}^{uc}}} f_{\text{GP}}(x^c, x^{uc}_{1}, x^{uc}_{2}, x^{uc}_{3}, x^{uc}_{4})
\end{equation}

\begin{align*}
\text{subject to} &~x^c = r , \text{(uniform flow rate for all blocks)}, \\
& x^{uc}_{k} = c_{k} , \text{(temperature setting for block $k$)}, \\
& k \in {1, 2, 3, 4} , \text{(indexing the blocks)}.
\end{align*}

The chosen design space is confined to flow rates within $[5 \, \text{ml/min}, 50 \, \text{ml/min}]$ and temperatures spanning $[520^\circ C, 590^\circ C]$ for each block \( k \).

This formulation sets the groundwork for the subsequent sections that delve into the specifics of the process-constrained parallel Bayesian Optimization (pc-BO) and its application to the Flowrence unit’s empirical data, aiming to demonstrate the practical efficacy of the pc-BO-TS method within a high-throughput setting.

%%%
\subsection{Results and Discussion}
\label{subsec:results_discussion_real}
This section presents the outcomes of optimization experiments conducted 
on the yield function \( f_{\text{ref}}(x) \) derived from real experimental data. 
The performance of various methods is quantified using the logarithm of normalized regret as a metric. 
These methods are consistent with those described in Section \ref{sec:syntheticTestCase}. 
Each method is run for ten identical random initializations in the design space, followed by 70 optimization iterations.

\begin{figure}%[h]
    \centering
    \includegraphics[width=0.8\textwidth]{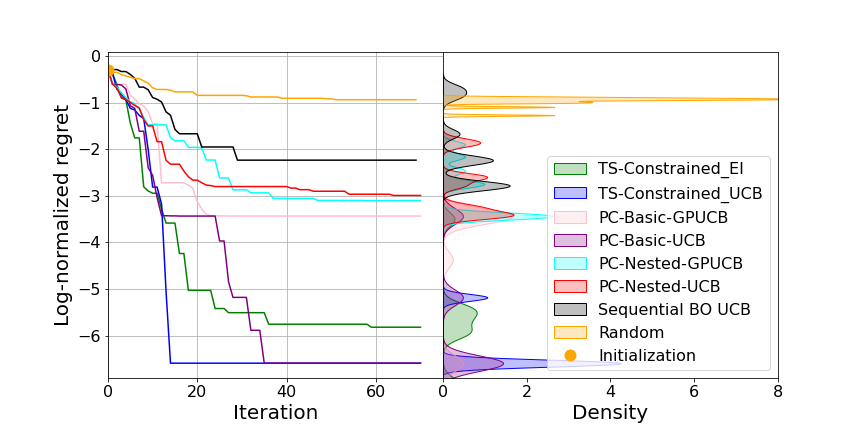}
    \caption{Optimization results based on a yield function based on experimental data. On the left, the log normalized median regret is visualized, and on the left one sees the final regret distribution. Both figures share the same y-axis.}
    \label{fig:experimental_test}
\end{figure}

%% Fig. 7
Figure \ref{fig:experimental_test} shows the results of these experiments.
Thompson sampling-based methods exhibit robust and quick asymptotic convergence. Specifically, TS-Constrained\_UCB demonstrates remarkable efficiency, achieving a 
log normalized regret as low as -6.6 after only 13 iterations. 
TS-Constrained\_EI also performs notably well, albeit converging to a slightly higher regret value of around -5.8 after 37 iterations. 
PC-Basic-UCB is worth mentioning as well, it attained a final regret level of -6.6. However, it it took 35 iterations to reach its final regret level. 
%
% Revision 2 Comment 1
The inclusion of the EI acquisition function alongside the UCB class further validates the robustness of the TS approach. TS-Constrained\_EI also performs remarkably well, illustrating that the benefits of our TS-based methods extend beyond a single class of acquisition functions.
An interesting observation is the similar rapid convergence pattern of TS-Constrained\_UCB, TS-Constrained\_EI, PC-Basic-UCB, 
and PC-Nested-GPUCB within the 12 first iterations. However, 
PC-Nested-GPUCB displayed a slower convergence rate in later stages, 
suggesting that its nested maximum variance approach and adaptive GPCUB strategy might lead 
to suboptimal exploration within the design space. 

A distinct aspect of the results is the stability exhibited by the pc-TS approaches in general in converging to the optimum, 
with TS-Constrained\_UCB being slightly less stable than TS-Constrained-EI. TS-Constrained-UCB shows two modes in the KDE in terms of the final regret, while TS-Constrained\_EI only shows one moderately wide mode. The pc-TS approaches with their more goal-oriented strategy show significantly quicker convergence compared to the more exploratory pc-BO(nested) and pc-BO(basic) approaches. It is worth mentioning that PC-Basic-UCB displays good asymptotic convergence. 
Considering that real experiments on the Flowrence unit are unlikely to exceed 
15 iterations, it is evident that TS-Constrained\_UCB, TS-Constrained\_EI, and PC-Basic-UCB are 
the most promising methods for achieving high accuracy within an experimental context. This finding 
underscores the potential of these three approaches in successfully optimizing the yield of a reaction within 
the constraints of the given setup.

%\clearpage
\section{Conclusion}
\label{sec:conclusion}

This research addresses the complex challenge of process-constrained batch optimization in high-throughput experimentation within a multi-reactor system. Traditional approaches, such as those proposed by \citet{Vellanki2017}, while offering theoretical and empirical improvements, often fall short in efficiency, especially under the stringent experimental constraints and cost considerations inherent to multi-reactor systems. % for heterogeneous catalytic HTE scenarios.

Many multi-reactor systems, including heterogeneous catalytic high throughput experimental scenarios, can be modeled as a hierarchy of embedded systems with different levels of control, see Fig.~\ref{fig:hierarchy_scheme}. Turning to the optimization of reaction conditions, this type of hierarchical experimental setup translates into a hierarchy of constrained/unconstrained parameters.
To tackle the corresponding optimization problem, we introduced the process-constrained Bayesian optimization with Thompson sampling (pc-BO-TS) as well as its hierarchical general version (hpc-BO-TS), a novel method tailored to overcome the limitations of previous works. By integrating Thompson sampling, our approach strikes a critical balance between exploration and exploitation under process constraints. 
An extensive set of numerical experiments, based on both synthetic and realistic objective functions, demonstrated that pc-BO-TS and hpc-BO-TS empirically surpass existing methods in terms of rapid convergence and robustness while taking into account the process constraints of MRS.
% Revision 2 Comment 1
Furthermore, those improvements are not limited to the UCB acquisition function but can be observed for EI as well, which confirms the general applicability of the proposed method.
These findings mark a contribution to the field of digital catalysis, showcasing a pragmatic application of advanced computational methods in multi-reactor experimental settings.

The primary contributions of this work lie in the development of a more efficient and effective optimization method for process-constrained batch problems; its successful application to realistic HTE scenarios illustrates the practical relevance of the proposed approach. The method’s adaptability in navigating complex hierarchical design spaces, while adhering to operational constraints, provides a valuable tool for researchers and practitioners in chemical engineering and related fields like digital catalysis.

%% Prospect
Looking ahead, several new perspectives appear. 
With process-constrained optimization methods at hand, the concept of multi-reactor systems could be thought of differently to prepare a more efficient search for optimal reaction conditions.
This leads to the concept of co-design, potentially shedding new light on experimental workflows in areas like heterogeneous catalysis: the experimental setup would be thought to optimize the performance of the later process-constrained optimization.
An important problem in this domain is the computational demand of Gaussian Processes (GPs), particularly in high-dimensional settings. Tackling this issue by adapting our approach, potentially in combination with sparse Gaussian Processes \citep{NIPS2005_sparse_GP_ghahramani}, presents an interesting research direction. Another direction for improvement may be the promotion of diversity within batches, taking inspiration from works on repulsive point process \citep{Kathuria2016BatchedGP}.

Such advancements could further enhance the efficiency and applicability of Bayesian optimization methods in complex, high-dimensional experimental designs, offering a ground for future exploration and development in machine learning and chemical engineering.

\section*{Credit authorship contribution statement}
\textbf{Markus Grimm}: Conceptualization, Experiments, Numerical experiments, Data curation, Formal analysis,
Investigation, Methodology, Software, Validation, Visualization,
Writing – original draft, Writing – review \& editing. \textbf{Sébastien Paul}:
Conceptualization, Data curation, Formal analysis, Supervision, Methodology, Project administration, Funding acquisition, Writing
– review editing, \textbf{Pierre Chainais}:
Conceptualization, Data curation, Formal analysis, Supervision, Methodology, Project administration, Funding acquisition, Writing
– review editing,

\section*{Declaration of competing interest}
The authors declare that they have no known competing financial interests or personal relationships that could have appeared to influence the work reported in this paper.

\section*{Data availability}
All the code i.e. numerical experiments, method implementations,
and scripts needed to reproduce the results and visualizations in this
paper can be found in the following GitHub repository: \href{https://github.com/Markusmu93/J-CompChE-HPCBoTS}{J-CompChE-HPCBoTS}.

\section*{Acknowledgements}
This project received funding from the European Union's Horizon 2020 research and innovation program under the Marie Skłodowska-Curie grant agreement No 847568.
The REALCAT platform is benefiting from a state subsidy administrated by the French National Research Agency (ANR) within the frame of the ‘Future Investments’ program (PIA), with the contractual reference ‘ANR-11-EQPX-0037’. The European Union, through the ERDF funding administered by the Hauts-de-France Region, has co-financed the platform. Centrale Lille, CNRS, and Lille University as well as the Centrale Initiatives Foundation, are thanked for their financial contributions to the acquisition and implementation of the equipment of the REALCAT platform.
This work was supported by the AI Sherlock Chair (ANR-20-CHIA-0031-01), the ULNE national future investment program (ANR-16-IDEX-0004) and the Hauts-de-France Region.

%%%%%%%%%%%%%%%%%%%%%%%%%%%%% APPENDIX %%%%%%%%%%%%%%%%%
\appendix

\section{Notation Table}
\label{sec:appendix_notation_table}

\begin{table}[h]
\centering
\caption{Updated Notation Table}
\label{tab:notation_updated}
\begin{tabular}{ll}
\hline
\textbf{Symbol} & \textbf{Description} \\
\hline
\multicolumn{2}{l}{\textbf{General Notation}} \\
\(f\) & 'Black box' objective function \\
\(\mathcal{X}\) & Domain of the objective function, subset of \(\mathbb{R}^n\) \\
\(\mathbf{x}^*\) & Optimal value that maximizes or minimizes \(f(\mathbf{x})\) \\
\(\mathbf{x}_t\) & Input to experiment at iteration \(t\) \\
\(R_T\) & Cumulative regret after \(T\) iterations \\
\(T\) & Total number of iterations \\
\(\mathcal{GP}(\mu, k)\) & Gaussian Process with mean function \(\mu\) and covariance function \(k\) \\
\(\alpha_{UCB}\) & Upper Confidence Bound acquisition function \\
\(D_t\) & Data observed up to iteration \(t\) \\
\(B\) & Batch size in Parallel Bayesian Optimization \\
\hline
\multicolumn{2}{l}{\textbf{Gaussian Process}} \\
\(\mu_{\theta}(\cdot) : \mathcal{X} \to \mathbb{R}\) & Mean function of the Gaussian Process, parameterized indirectly by \(\theta\) \\
\(k_{\theta}(\cdot, \cdot) : \mathcal{X} \times \mathcal{X} \to \mathbb{R}\) & Covariance function (kernel) of GP, parameterized by \(\theta\) \\
\(\theta\) & Set of hyperparameters for the covariance function \\
\(\boldsymbol{\varphi}\) & Function values of \(f\) at points \(x_{1:n}\), prior to observation noise \\
\(\boldsymbol{X}\) & Set of input points \\
\(\boldsymbol{y}\) & Noisy observations corresponding to inputs \(\boldsymbol{X}\) \\
\(\sigma^2\) & Variance of observation noise \\
\(\mathcal{N}(\boldsymbol{m}_{\theta}, \boldsymbol{K}_{\theta})\) & Multivariate normal distribution with mean \(\boldsymbol{m}_{\theta}\) and covariance \(\boldsymbol{K}_{\theta}\) \\
\(\mathcal{N}(\boldsymbol{f}, \sigma^2\mathbf{I})\) & Normal distribution of observations given function values \(\boldsymbol{f}\) and variance \(\sigma^2\) \\
\(\mu_{t, \theta}(x)\) & Posterior mean of GP at point \(x\) and time \(t\), parameterized indirectly by \(\theta\) \\
\(k_{t, \theta}(x, x')\) & Posterior covariance between points \(x\) and \(x'\) at time \(t\), parameterized by \(\theta\) \\
\(\sigma_{\theta,t}^2(x)\) & Posterior variance of GP at point \(x\) and time \(t\), parameterized by \(\theta\) \\
\(\boldsymbol{K}_{\theta,t}\) & Covariance matrix at time \(t\), parameterized by \(\theta\)\\
\(\boldsymbol{m}_{t, \theta}\) & Mean vector at time \(t\), indirectly parameterized by \(\theta\)\\
\(\boldsymbol{C}_{t, \theta}\) & Covariance matrix at time \(t\), parameterized by \(\theta\), including noise term \(\sigma^2\mathbf{I}_t\) \\
\hline
\multicolumn{2}{l}{\textbf{Bayesian Optimization}} \\
\( \alpha_{\text{GP-UCB}}^{(t)}(x) \) & GP-UCB acquisition function at time \( t \) and point \( x \) \\
\( \beta_t \) & Confidence parameter in GP-UCB \\
\( \mu_{t-1}(x) \) & Posterior mean of GP at point \( x \) and time \( t-1 \) \\
\( \sigma_{t-1}(x) \) & Posterior standard deviation of GP at point \( x \) and time \( t-1 \) \\
\( \alpha_{\text{EI}}^{(t)}(x) \) & Expected Improvement acquisition function at time \( t \) and point \( x \) \\
\( f(x^*_{t-1}) \) & Best observed objective function value up to iteration \( t-1 \) \\
\( \Phi(z) \) & Cumulative distribution function of the standard normal distribution \\
\( \phi(z) \) & Probability density function of the standard normal distribution \\
\( z \) & Standardized variable in Expected Improvement function \\
\( D_t \) & Cumulative current query-observation pairs up to iteration \( t \) \\
\hline
\multicolumn{2}{l}{\textbf{Process-Constrained Batch Bayesian Optimization (pc-BO)}} \\
\( \alpha_{UCB}(\boldsymbol{x}_{t,0} | D_{t-1}) \) & UCB acquisition function for pc-BO at iteration \( t \) \\
\( \mathbf{x}^{uc} \) & Unconstrained variables \\
\( \mathbf{x}^{c} \) & Constrained variables \\
\( X^c \) & Space of constrained parameters \\
\( X^{uc} \) & Space of unconstrained parameters \\

\hline
\end{tabular}
\end{table}

%\clearpage

%%%%%%%%%
\section{Implementation Details}
A key set of implementation details about the methods used and the benchmarks are included in this section. For full details, we refer any reader to the code, available at \href{https://github.com/Markusmu93/J-CompChE-HPCBoTS}{J-CompChE-HPCBoTS}.

\subsection{Process constrained batch BO methods}
Besides our method pc-BO-TS, we implemented pc-BO-Nested, and pc-BO-Basic ourselves since Vellanki et. al. did not provide any implementation. We utilize a Gaussian Process (GP) model from scikit-learn, incorporating the Matern kernel:

\begin{equation}
k_{\text{Matern}}(x_1, x_2) = \theta_0 \frac{2^{1-\nu}}{\Gamma(\nu)}\left(\sqrt{2\nu}\frac{d(x_1, x_2)}{\ell}\right)^\nu K_\nu\left(\sqrt{2\nu}\frac{d(x_1, x_2)}{\ell}\right)
\end{equation}
where $d(x_1, x_2)$ is the Euclidean distance between points $x_1$ and $x_2$, $\theta_0$ is the output scale, $\ell$ is the length-scale, $\nu$ is the smoothness parameter (set to 2.5), and $K_\nu$ is a modified Bessel function \citep{Rasmussen2004}.

Hyperparameters of these GP models are updated at each iteration by maximizing the marginal log-likelihood. The optimization in all methods starts with a single, randomly chosen point in the design space. The DIRECT algorithm is employed for optimizing all acquisition functions presented in this work such as GP-UCB or GP-UCB-PE.

\textit{pc-BO-TS} differentiates between constrained and unconstrained variables in the design space exploration. For unconstrained variables, it constructs a regular grid of $10^d$ points, where $d$ is the dimensionality of the unconstrained space. The algorithm then evaluates function samples at these grid points

\subsection{Synthetic test cases}
\label{subsec:synthetic-functions}
In the following, we provide all the information of  test functions used in our work regarding its form and the specific domain used: 

\textbf{Levy 6D} \\
\begin{equation}
L(\mathbf{x}) = 47.341 - (\sin^2(\pi w_1) + \sum_{i=1}^{5} \left[(w_i - 1)^2 (1 + 10 \sin^2(\pi w_i + 1))\right] + (w_6 - 1)^2 (1 + \sin^2(2 \pi w_6)))
\end{equation}
where \( \mathbf{x} = [x_1, x_2, ..., x_6] \) is a 6-dimensional input vector and \( w_i = 1 + \frac{x_i - 1}{4} \) for \( i = 1, 2, ..., 6 \).

The domain for this function is defined as \(\mathcal{X} = \mathcal{X}^c \times \mathcal{X}^{uc}\), with:
\begin{align*}
    \mathcal{X}^c &= \{ x \in \mathbb{R}^3 : -5 \leq x_i \leq 5, \; i = 1, 2, 3 \}, \\
    \mathcal{X}^{uc} &= \{ x \in \mathbb{R}^3 : -5 \leq x_j \leq 5, \; j = 4, 5, 6 \}.
\end{align*}
Here, \(\mathcal{X}^c\) represents the constrained subspace for dimensions \(x_1, x_2, x_3\), and \(\mathcal{X}^{uc}\) the unconstrained subspace for dimensions \(x_4, x_5, x_6\), all within the range \([-10,10]\). 

\textbf{Hartmann 6D} \\
\begin{equation}
H(\mathbf{x}) = \sum_{i=1}^{4} \alpha_i \exp \left( -\sum_{j=1}^{6} A_{ij} (x_j - P_{ij})^2 \right)
\end{equation}
where \( \mathbf{x} = [x_1, x_2, ..., x_6] \) is a 6-dimensional input vector, \( \alpha = [1.0, 1.2, 3.0, 3.2] \), and matrices \( A \) and \( P \) are given by:
\[
A = \begin{pmatrix}
    10 & 3 & 17 & 3.5 & 1.7 & 8 \\
    0.05 & 10 & 17 & 0.1 & 8 & 14 \\
    3 & 3.5 & 1.7 & 10 & 17 & 8 \\
    17 & 8 & 0.05 & 10 & 0.1 & 14
\end{pmatrix}, \quad
P = 10^{-4} \times \begin{pmatrix}
    1312 & 1696 & 5569 & 124 & 8283 & 5886 \\
    2329 & 4135 & 8307 & 3736 & 1004 & 9991 \\
    2348 & 1451 & 3522 & 2883 & 3047 & 6650 \\
    4047 & 8828 & 8732 & 5743 & 1091 & 381
\end{pmatrix}
\]
The domain for this function is defined as \(\mathcal{X} = \mathcal{X}^c \times \mathcal{X}^{uc}\), with:
\begin{align*}
    \mathcal{X}^c &= \{ x \in \mathbb{R}^3 : 0 \leq x_i \leq 1, \; i = 1, 2, 3 \}, \\
    \mathcal{X}^{uc} &= \{ x \in \mathbb{R}^3 : 0 \leq x_j \leq 1, \; j = 4, 5, 6 \}.
\end{align*}
Here, \(\mathcal{X}^c\) represents the constrained subspace for dimensions \(x_1, x_2, x_3\), and \(\mathcal{X}^{uc}\) the unconstrained subspace for dimensions \(x_4, x_5, x_6\), all within the range [0,1].

\textbf{Rosenbrock 3D} \\
The modified Rosenbrock 3D function is defined as:
\begin{equation}
  f(\mathbf{x}) = 7218-\Bigg[ (1 - x_1)^2 + 100 (x_2 - x_1^2)^2
                  + (1 - x_2)^2 + 100 (x_3 - x_2^2)^2 \Bigg],  
\end{equation}
over $X = [-2, 2]^3$. The negative sign transforms the traditional minimization problem into a maximization problem. 
The global maximum value of the function is \(f(\mathbf{x}^*) = 7218\) at the 
point \(\mathbf{x}^* = (1, 1, 1)\), and the minimum value is 0 over $[-2, 2]^3$, reached in $(-2,-2,-2)$.%\\[3mm]

\textbf{Rosenbrock 4D} \\
The modified 4D Rosenbrock function \( R(\mathbf{x}) \) is defined as:
\begin{equation}
R(\mathbf{x}) = 10827 -\sum_{i=1}^{3} \left[100 (x_{i+1} - x_i^2)^2 + (1 - x_i)^2\right]
\label{eq:rosenbrock_app_4d}
\end{equation}
where \( \mathbf{x} = [x_1, x_2, x_3, x_4] \) is a 4-dimensional input vector.

The domain configurations for the test cases are defined for \( k \) constrained dimensions (\( k = 1, 2, 3 \)) as follows:
\begin{equation}
\mathcal{X}_k = \mathcal{X}^{uc} \times \mathcal{X}^c
\end{equation}
where
\begin{align*}
    \mathcal{X}^{uc} &= \{ x \in \mathbb{R}^{4-k} : -2 \leq x_i \leq 2, \; i = 1, \ldots, 4-k \}, \\
    \mathcal{X}^c &= \{ x \in \mathbb{R}^{k} : -2 \leq x_i \leq 2, \; i = 4-k+1, \ldots, 4 \}.
\end{align*}

\subsection{Realistic Test Case}
\label{subsec:realistic_test_case_appendix}
For our surrogate model of the ODHP yield function, we employed a GP model with a Radial Basis Function (RBF) kernel, as implemented in scikit-learn. The GP model is mathematically represented as:

\begin{equation}
f_{\text{GP}}(x) \sim \mathcal{GP}(m(x), k_{\text{RBF}}(x, x')),
\end{equation}

where \( f_{\text{GP}}(x) \) denotes the Gaussian Process, encapsulating a distribution over functions. The mean function \( m(x) \) is typically set to zero, and \( k_{\text{RBF}}(x, x') \) is the RBF kernel, expressed as:

\begin{equation}
k_{\text{RBF}}(x, x') = \exp\left(-\frac{\|x - x'\|^2}{2l^2}\right),
\end{equation}

with \( \|x - x'\|^2 \) representing the squared Euclidean distance between inputs \( x \) and \( x' \), and \( l \) being the length scale parameter.

In practice, \( f_{\text{GP}}(x) \) is used to predict the yield at new input points \( X_{\text{new}} \). These predictions are computed as the posterior mean of the GP process, given the observed data:

\begin{equation}
\hat{y} = \mathbb{E}[f_{\text{GP}}(X_{\text{new}}) | X, y].
\end{equation}

The training of the GP model involved preparing input-output pairs \((X, y)\) from the experimental dataset. These pairs consisted of the input variables (flow rates, temperatures) and the corresponding yield measurements. The hyperparameters of the GP model, including the length scale \(l\) of the RBF kernel, were then optimized to maximize the model's fit to this data via log marginal likelihood. Following this, the GP model was trained with the dataset \((X, y)\), enabling it to learn the relationship between the experimental conditions and the yield. $f_{GP}$ can be found as a pickled scikit-learn gp regression model in \href{https://github.com/Markusmu93/J-CompChE-HPCBoTS}{J-CompChE-HPCBoTS}.

 \bibliographystyle{elsarticle-harv} 
 \bibliography{cas-refs}

\end{document}